\documentclass[lettersize,journal]{IEEEtran}
\usepackage{amsmath,amsfonts}
\usepackage{array}
\usepackage[caption=false,font=normalsize]{subfig}
\usepackage{textcomp}
\usepackage{stfloats}
\usepackage{url}
\usepackage{verbatim}
\usepackage{graphicx}
\usepackage{cite}

\usepackage{array}
\usepackage{amsmath}
\usepackage[linesnumbered,ruled,vlined]{algorithm2e}
\usepackage{amssymb}
\usepackage{multirow}
\usepackage{booktabs}
\usepackage{hyperref}
\usepackage{dcolumn}
\usepackage{siunitx}

\mathchardef\mhyphen="2D

\usepackage{cleveref}
    \crefformat{section}{Sec.~#2#1#3}
    \crefformat{figure}{Fig.~#2#1#3}
    \crefformat{table}{Table~#2#1#3}
    \crefformat{equation}{Eq.~#2#1#3}
    \crefformat{algorithm}{Alg.~#2#1#3}

\makeatletter
\algocf@newcommand{InputXX}[1]{%
  \sbox\algocf@inputbox{\hbox{\KwSty{Output}\algocf@typo: }}%
  \ifthenelse{\boolean{algocf@inoutnumbered}}{\relax}{\everypar={\relax}}%
  {\let\\\algocf@newinput\hspace{\wd\algocf@inputbox}\hangindent=\wd\algocf@inputbox\hangafter=\wd\algocf@inputbox#1\par}%
  \algocf@linesnumbered% reset the numbering of the lines
}
\makeatother

\def\appendixautorefname~#1\null{~#1 \null}
\def\algorithmautorefname~#1\null{Alg.~#1\null}

\usepackage{empheq}
\usepackage[most]{tcolorbox}
\tcbset{
    colorbox/.style={
        enhanced,
        frame hidden,
        colback=#1,         % 背景色を変数に
        top=0pt,
        bottom=0pt,
        left=0pt,
        right=0pt
    },
    colorbox/.default=red!10
}

\usepackage{xspace}
\makeatletter
\DeclareRobustCommand\onedot{\futurelet\@let@token\@onedot}
\def\@onedot{\ifx\@let@token.\else.\null\fi\xspace}

\def\ie{\emph{i.e}\onedot}

\def\etal{\emph{et al}\onedot}
\makeatother
\graphicspath{{./imgs/}}

\begin{document}

\let\ref\cref

\title{Weakly-Supervised Crack Detection}

\author{Yuki Inoue and Hiroto Nagayoshi
        % <-this % stops a space
\thanks{Y. Inoue and H. Nagayoshi are with the Center of Artificial Intelligence, Hitachi Ltd, Tokyo, Japan, (e-mail: yuki.inoue.wh@hitachi.com, hiroto.nagayoshi.wy@hitachi.com).}
% <-this % stops a space
\thanks{This work has been submitted to the IEEE for possible publication. Copyright may be transferred without notice, after which this version may no longer be accessible.}}


\maketitle

\begin{abstract}
    Pixel-level crack segmentation is widely studied due to its high impact on building and road inspections.
    While recent studies have made significant improvements in accuracy, they typically heavily depend on pixel-level crack annotations, which are time-consuming to obtain.
    In earlier work, we proposed to reduce the annotation cost bottleneck by reformulating the crack segmentation problem as a weakly-supervised problem- \ie the annotation process is expedited by sacrificing the annotation quality.
    The loss in annotation quality was remedied by refining the inference with per-pixel brightness values, which was effective when the pixel brightness distribution between cracks and non-cracks are well separated, but struggled greatly for lighter-colored cracks as well as non-crack targets in which the brightness distribution is less articulated.
    In this work, we propose an annotation refinement approach which takes advantage of the fact that the regions falsely annotated as cracks have similar local visual features as the background. Because the proposed approach is data-driven, it is effective regardless of a dataset's pixel brightness profile.
    The proposed method is evaluated on three crack segmentation datasets as well as one blood vessel segmentation dataset to test for domain robustness, and the results show that it speeds up the annotation process by factors of 10 to 30, while the detection accuracy stays at a comparable level.
\end{abstract}


\section{Introduction} \label{sec:intro}

\IEEEPARstart{K}{eeping} roads crack-free is crucial for public safety, as cracks are the first signs of structural deterioration, and they may lead to more severe problems such as potholes and collapses. As such, automatic crack detection is critical for ensuring safety in public transportation systems.

Crack detection problem is typically formulated as a semantic segmentation problem, as it is crucial to gather information about various crack properties such as width and orientation in addition to location to accurately assess the conditions of the target structure \cite{inoue, yang2019feature, liu2019deepcrack, guo2021barnet, qu2021crack}. Such analysis is only possible with the details provided by the segmentation outputs.
% Such analysis is impossible with other problem settings like object detection and image classification.
However, one major bottleneck with semantic segmentation is the annotation cost, as pixel-level annotation is one of the most cost-intensive annotations to obtain.
There are two reasons why pixel-level annotations are especially problematic for crack segmentation problem.
First, cracks can be arbitrarily thin; they can be as thin as one pixel or even sub-pixel. As a result, there is little to no tolerance for annotation errors, making crack annotations even more expensive than typical segmentation annotations.
Second, cracks can form on many different structures such as roads, bridges, and buildings, and the appearance of a crack varies greatly across surrounding environments and materials of the target structure, making it difficult to maintain a single universal crack detection model that can be used anywhere. Therefore, it is recommended to prepare new sets of annotations at each site, implying that the annotation cost is recurring in a crack detection service.
Unfortunately, only a few studies focus on annotation-efficient crack detection.
We believe that performance improvement and annotation reduction strategies should be studied in parallel so that in the future, accurate crack detection systems can be provided at low costs.

In the earlier version of this manuscript \cite{inoue2020crack}, we proposed to use imprecise annotations during model training, \ie formulate the problem as a weakly-supervised problem.
While imprecise annotations significantly reduced the preparation cost, they lead to imprecise model training. In order to counteract the negative effects of the low quality annotations, we introduced a two-branch detection framework. In the proposed framework, we complement the conventional data-driven inference model (named \textit{Macro Branch}) with a rule-based inference path (named \textit{Micro Branch}), which bases its decision solely on per-pixel darkness. The idea is based on the observation that cracks are typically darker than their surroundings.
As such, the proposed method performed well when the pixel brightness distribution between cracks and non-cracks are well separated. In fact, the method worked so well that the accuracy performance at lower annotation quality matched or even exceeded the performance for the case in which accurate annotation is used; one can say that weakly-supervised problem is solved when cracks are significantly darker than their surroundings. However, the method struggled greatly for the case in which crack and non-crack pixels have similar brightness distributions, and it was left as a future challenge.

Recognizing these problems with our earlier work, this version makes several new contributions.
First, we introduce a new module that refines the annotations used to train the Macro Branch. We take advantage of the fact that the regions falsely annotated as cracks (\ie actually non-cracks but annotated as cracks because of the weak annotation settings) have similar local visual features as the background region (\ie actually non-cracks and annotated as non-cracks). Under such a situation, the two regions produce conflicting gradient updates, but the wrong gradient updates from the mis-annotated regions get dominated by the correct gradient updates from the background region. This occurs because the mis-annotation only occurs near the crack regions, and therefore the mis-annotated region is significantly smaller than the background region. We utilize this phenomenon to refine the annotations.
Second, since annotation refinement is data-driven, we hypothesize that it is effective even for the cases in which the pixel darkness-based Micro Branch struggled, and confirm this hypothesis through evaluations on various crack segmentation datasets as well as a retinal blood vessel segmentation dataset.
Finally, the proposed method is compared against other annotation-reducing methods to show its superiority.

\section{Related Work}

Most recent works in the field of crack detection assume fully-supervised training with pixel-accurate annotations. However, some works have started to propose more annotation-efficient methods, recognizing the importance of reducing the annotation bottlenecks.
In this section, we will review the previous strategies for reducing annotation costs, categorized by the type of annotations.

\subsection{Annotation-Free} \label{ssec:annotation_free}

Taking annotation reduction efforts to the extreme, we have the annotation-free approaches, sometimes referred to as unsupervised training, in which the training is performed without any annotations.
In a way, rule-based crack detectors can be thought of as unsupervised methods as they do not require explicit annotations. Unfortunately, they often require experts to tune the parameters when adapting to new environments, which negates the benefit of being annotation-free. In addition, rule-based approaches are known to be inferior in performance when compared to data-driven counterparts \cite{yang2019feature, liu2019deepcrack}. 

Another form of annotation-free approach is to train a model on a separate dataset and apply it in the new environment, with no fine-tuning. This is the ``universal crack detection model'' mentioned in \mbox{\ref{sec:intro}}, and will be referred to as the out-of-domain (\textit{OOD}) case. Though OOD is not explicitly studied in any literature, some have adopted it in the model evaluation, in which the models are trained and tested on different datasets \mbox{\cite{konig2021weakly, yang2019feature}}. Unfortunately, there is a significant drop in performance compared to the case in which the train and test datasets match (in-domain case, \textit{ID}).

More recently, data-driven methods such as ones based on color histograms \cite{mubashshira2020unsupervised} and Generative Adversarial Networks \cite{duan2020unsupervised, yu2020unsupervised} are developed for unsupervised crack detection. In many of these methods, models learn to reconstruct non-defective images. Crack detection is achieved through the model's inability to accurately reconstruct the anomaly images.
However, they are still inaccurate compared to their supervised counterparts. 
For example, Duan \etal and Mubashshira \etal \cite{mubashshira2020unsupervised, duan2020unsupervised} either perform marginally better than rule-based methods or worse than the supervised methods.
On the other hand, Yu \etal \cite{yu2020unsupervised} claim that their unsupervised model outperforms supervised models for almost all datasets. However, their claim has two problems. First, out of the supervised models compared in the paper, only FPHBN \cite{yang2019feature} can be considered competitive by today's standards. Furthermore, the performance statistics of FPHBN are taken from the original paper, which obtains the result by training FPHBN with CRACK500 and evaluating it across different datasets. In other words, they evaluate under OOD settings except for CRACK500. Therefore, the only valid comparison between supervised and unsupervised models is against FPHBN evaluated with CRACK500, in which Yu \etal's unsupervised model underperforms, breaking their claim.
Drops in performance can lead to cracks being overlooked, which in turn may lead to serious accidents. So if the small cost of creating rough annotations can significantly improve performance, it should be strongly preferred.

Aside from their segmentation performance, one main issue with most data-driven unsupervised methods is that the models are never explicitly taught what a crack looks like. They are instead only shown what a \textit{normal} scenery looks like, and taught to detect any unforeseen differences.
This means that if we want to ignore certain defect types, their samples must be included in the training dataset so that the models learn to recognize those defects as a part of a normal scene. For example, if a model should not detect rust, images of rusted objects must be present in the training dataset in a sizeable volume. This targeted sample collection adds hidden costs to the data preparation pipeline, potentially becoming more time-consuming than the annotation process and negating the benefits of the unsupervised approach.

\begin{table}
	\centering
	\caption{Dataset information and annotation times. Note that only CFD was re-annotated precisely for time measurement. \textit{Precise}, \textit{Rough}, and \textit{Rougher} correspond to types of annotations, details described in \ref{ssec:annotation}.}
	\begin{tabular}{@{\extracolsep{4pt}}cccccc@{}}
		\toprule
		\multirow{2}{*}{Dataset}& \multicolumn{2}{c}{Sample counts} & \multicolumn{3}{c}{Annotation time per image (sec.)} \\
		\cmidrule{2-3} \cmidrule{4-6}
                &Train  &Test   &Precise& Rough & Rougher \\
        \midrule
		Aigle   & 24    & 14    & -     & 34    & 23 \\
		CFD     & 71    & 47    & 656   & 70    & 22 \\
		DCD     & 300   & 237   & -     & 97    & 17  \\
		\bottomrule
	\end{tabular}
	\label{tbl:dataset_overview}
\end{table}

\begin{figure}[!tb]
    \footnotesize
    \centering
    \renewcommand{\arraystretch}{0.6}
    \setlength{\tabcolsep}{1pt}
    \newcolumntype{C}{>{\centering\arraybackslash} m{1.7cm} }
	\begin{tabular}{CCCCC}

    \includegraphics[width=\linewidth]{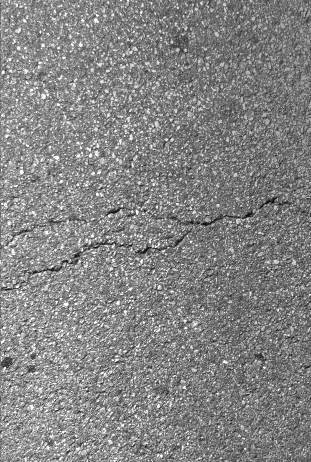} &
    \includegraphics[width=\linewidth]{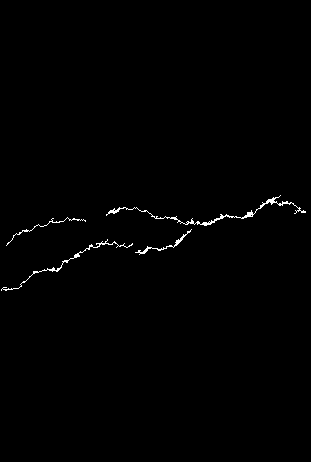} &
    \includegraphics[width=\linewidth]{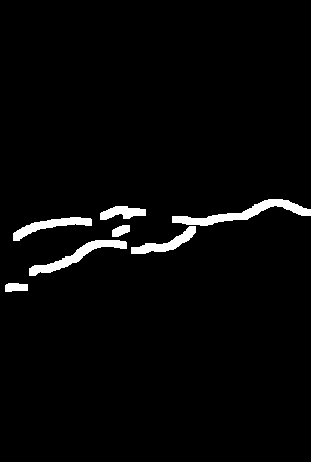} &
    \includegraphics[width=\linewidth]{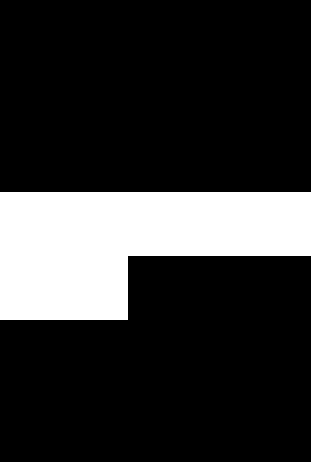} &
    \includegraphics[width=\linewidth]{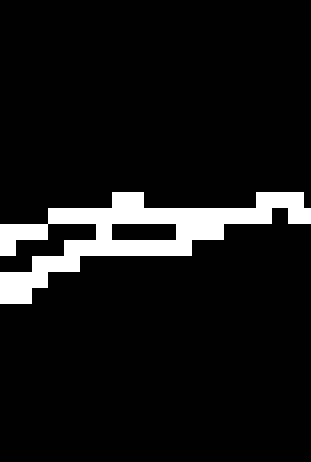} \\

    \includegraphics[width=\linewidth]{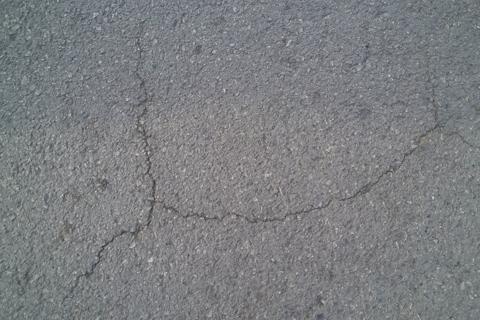} &
    \includegraphics[width=\linewidth]{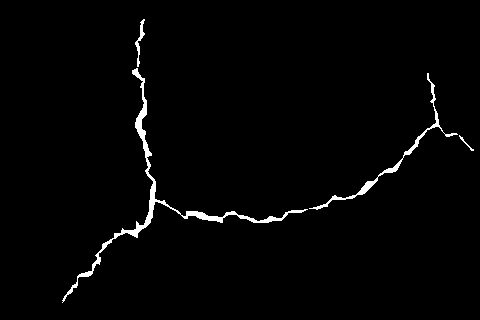} &
    \includegraphics[width=\linewidth]{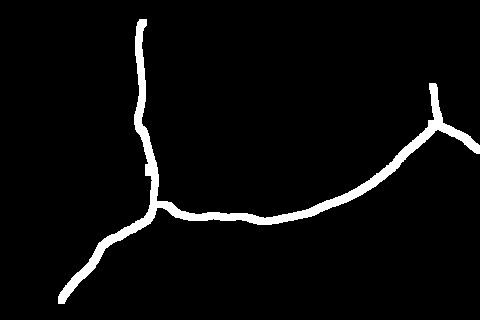} &
    \includegraphics[width=\linewidth]{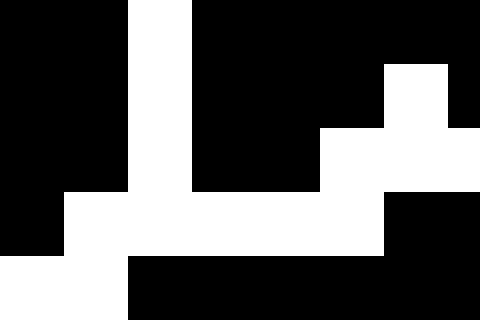} &
    \includegraphics[width=\linewidth]{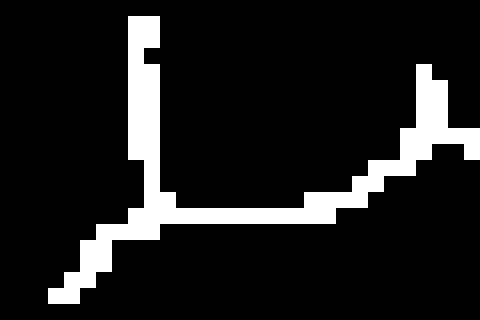} \\
    
    \includegraphics[width=\linewidth]{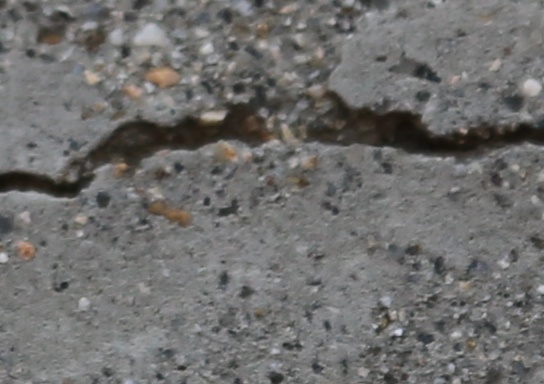} &
    \includegraphics[width=\linewidth]{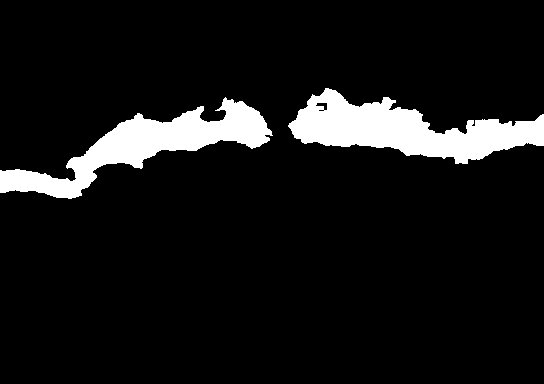} &
    \includegraphics[width=\linewidth]{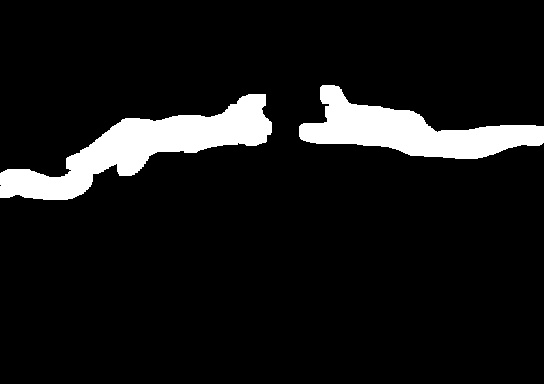} &
    \includegraphics[width=\linewidth]{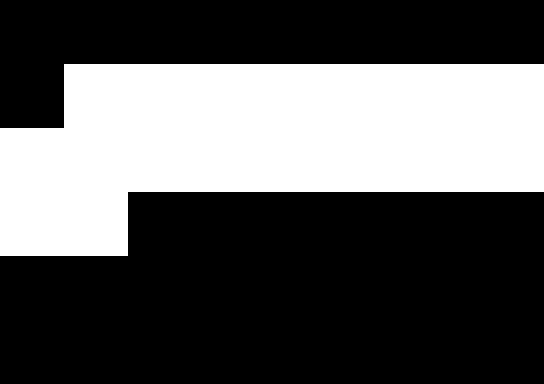} &
    \includegraphics[width=\linewidth]{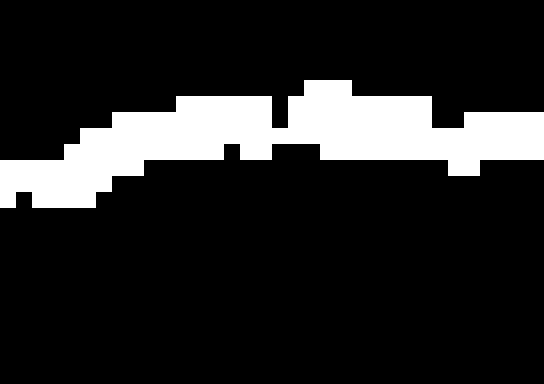} \\

    \\[-1.5mm]
    (a) Original & (b) Precise & (c) Pixel-level (Rougher Annotation) & (d) Patch-level ($64\times64$) & (e) Patch-level ($16\times16$) \\

    \end{tabular}
    \caption{Comparisons between the two types of weak supervision annotations. The white region represents the annotated crack region.}
    \label{fig:weak_annot_comp}
\end{figure}

\begin{figure}[!tb]
    \captionsetup[subfloat]{margin=4pt,format=hang,singlelinecheck=false,font=footnotesize}
    \centering

	\subfloat[Original image.]{
	    \includegraphics[width=0.3\linewidth]{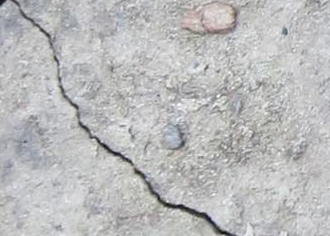}
	    \label{fig:two_weak_ori}
	}
	\subfloat[Patch-level.]{
	    \includegraphics[width=0.3\linewidth]{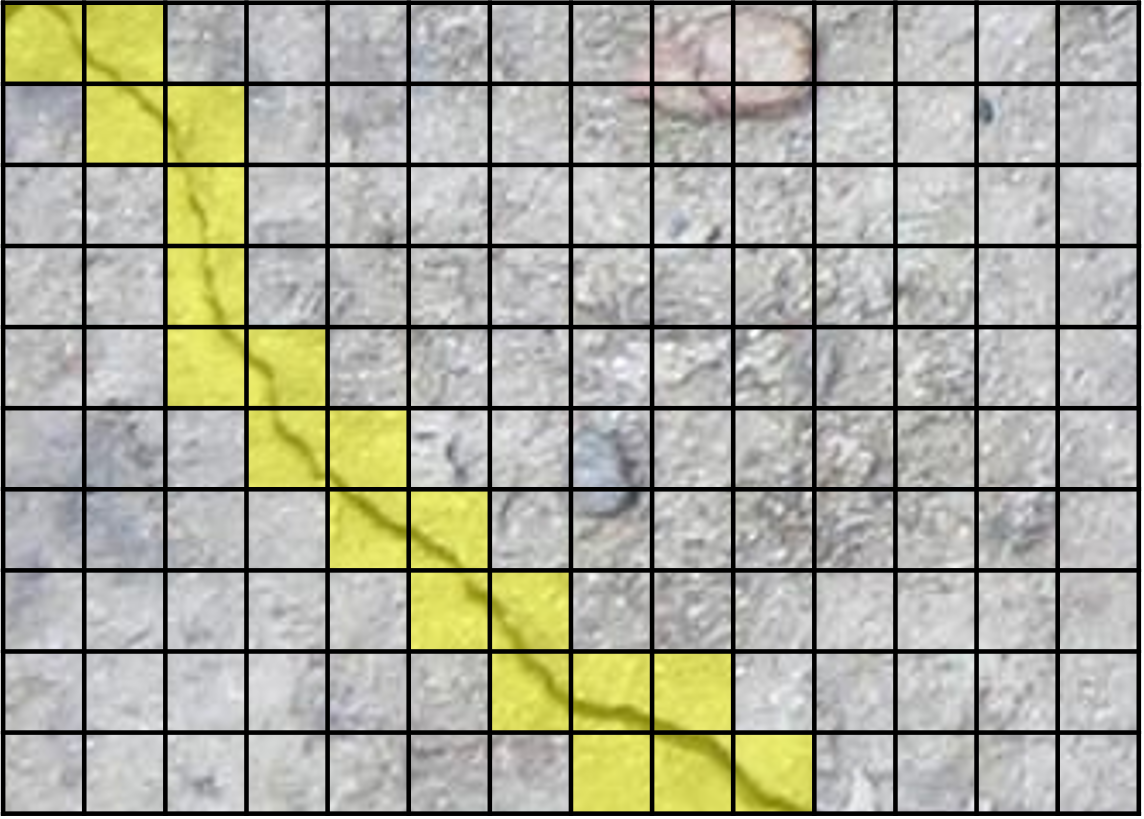}
	    \label{fig:two_weak_patch}
	}
	\subfloat[Pixel-level.]{
	    \includegraphics[width=0.3\linewidth]{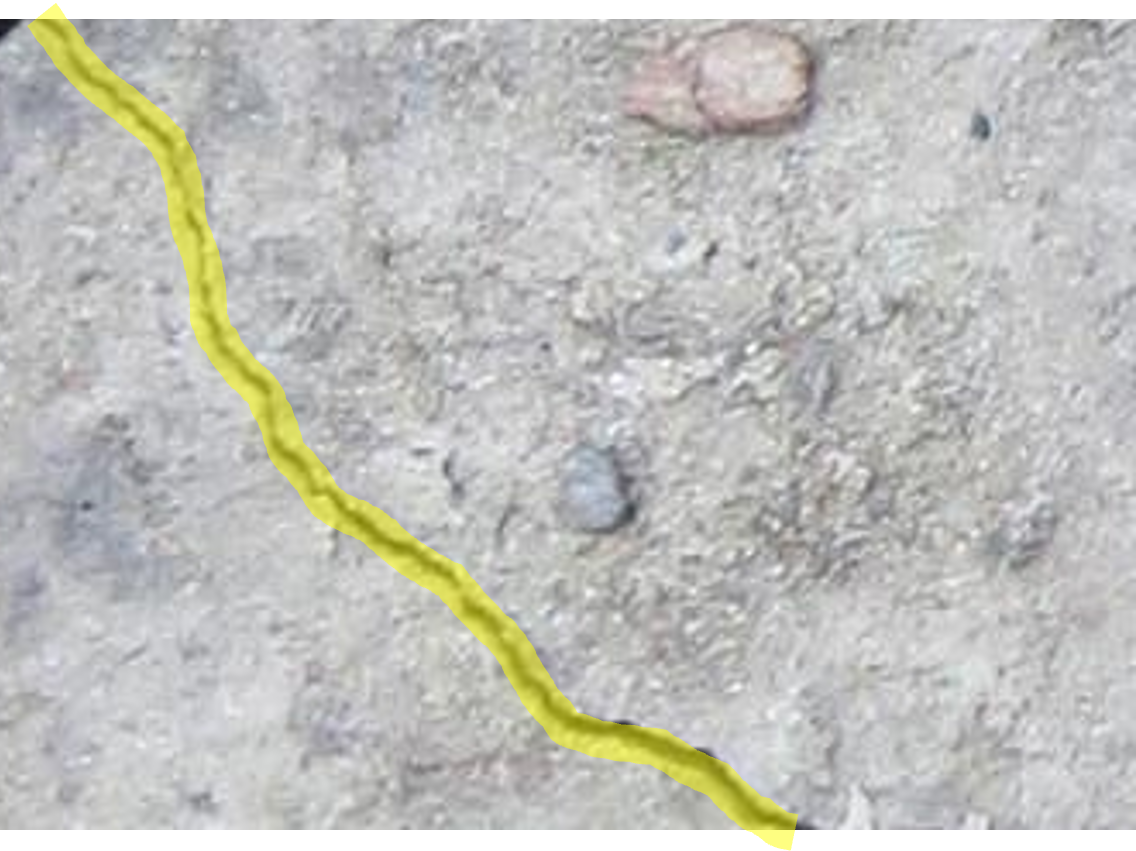}
	    \label{fig:two_weak_pixel}
	}

	\caption{An illustration of how an image may be annotated under different weak annotation strategies. Yellow pixels represent the regions annotated as cracks. In both cases, an annotator likely follows a similar annotation path (from the top of the image to the bottom of the image or vice versa), but our proposed pixel-level annotation produces a more natural crack annotation.}
	\label{fig:two_weak_sup}
\end{figure}

\subsection{Approximate Annotations} \label{ssec:annotation_weak}

Annotation costs can be reduced by compromising the annotation quality for speed.
Preparing approximate annotations that ignore the crack boundaries such as shown in \ref{fig:weak_annot_comp}c-e significantly speeds up the annotation process, because the boundaries between crack and non-crack regions are often blurry and ambiguous and thus take a long time to judge. In addition, even after the boundaries are determined, they are tedious to annotate, because boundaries are often complex to trace. The annotators can ignore both of these difficulties with approximate annotations and save much time.

However, approximate annotations are not without consequences- the models trained with such annotations will produce low quality outputs. This is a problem setting known as the weakly-supervised problem, and in the field of crack segmentation, two approaches have been proposed, differing in the format of the annotation.

First provides the supervision in a form of classification labels- cracks are subdivided into patches, and each patch is annotated whether it contains any cracks. \ref{fig:weak_annot_comp}d,e show examples of such annotations at two patch sizes of 64 and 16. We will refer to this as \textit{patch-level weak supervision}.
Fan \etal \cite{fan2019road} was one of the first to try this approach, training a classifier model with weak labels and applying a rule-based thresholding method for patches in which the trained classifier predicted as containing cracks. One major disadvantage of their proposal is that the supervised model can only isolate the crack regions up to square patches. Because cracks are thin, these selected patches mostly contain non-crack regions and are difficult to be refined.
K\"{o}nig \etal \cite{konig2021weakly} and Dong \etal \cite{dong2020patch} improved the approach by refining the patch-level estimation of the classifier with class activation maps (CAM) \cite{selvaraju2017grad}. The CAM outputs are further refined by thresholding methods such as conditional random field or Otsu's binarization.

We proposed in \cite{inoue2020crack} another form of weak supervision, \textit{pixel-level weak supervision}, which is a free-form imprecise semantic segmentation annotation as illustrated in \ref{fig:weak_annot_comp}c. Unlike patch-level weak supervision, it has much smoother outer edges that better trace the cracks. \ref{tbl:dataset_overview} summarizes the times spent on the annotation. \textit{Rough} and \textit{Rougher} correspond to the proposed annotation strategy (details explained in \ref{ssec:annotation}), and \textit{Precise} correspond to pixel-precise annotation. Since the annotation time for the pixel-precise annotation is not available, one of the datasets was re-annotated for the time-measuring purpose only. As the table shows, up to 96\% of the annotation time was saved with this annotation strategy.

One common assumption between the two annotation approaches is that the cracks are over-annotated. Between the true annotation and the weak annotation, very few are false-negatives (\ie true crack pixels annotated as non-cracks).

In this paper, the crack detection problem is formulated as a pixel-level weakly-supervised problem, as we believe it is more efficient than patch-level annotations.
Unfortunately, because procedures for generating patch-level annotations by hand are not discussed in literatures as they are automatically generated from the pixel-precise annotations that come with the datasets, their time efficiency can only be discussed qualitatively. Probably the most efficient method for annotating patches by hand is to overlay a grid of patch boundaries on the image as illustrated in \ref{fig:two_weak_patch}, and ask the annotators to select patches with cracks.
Since cracks are thin and connected, this selection process requires the annotators to trace each crack from start to finish. This is very similar to how annotation is done for pixel-level weak supervision, except that cracks are annotated in free-form instead of in grid-form, as shown in \ref{fig:two_weak_pixel} and \ref{fig:two_weak_patch}, respectively. As the annotation process is very similar to each other, we assume that their time efficiencies are also similar, especially for smaller patch sizes.

However, one key difference is that the patch-level weak annotations tend to be noisier. As they are built on a grid of patches, they can be interpreted as a spatially-discretized version of pixel-level annotations. \ref{fig:two_weak_sup} shows an example of the two annotations for the same crack, and we can see that the patch-level annotation lacks details compared to the pixel-level counterpart.
As a result, we conclude that pixel-level weak supervision has a better annotation cost to quality tradeoff.

\subsection{Pixel-Precise Annotations} \label{ssec:annotation_precise}

One way to reduce the annotation cost is to only annotate a subset of a dataset.
Li \mbox{\etal} and Shim \mbox{\etal} propose semi-supervised crack detection methods that train the models in an adversarial manner \mbox{\cite{li2020semi, shim2020multiscale}}. They train a discriminator network alongside a segmentation network, to be used to generate pseudo-labels for the unlabeled samples. The human-labeled annotations and the pseudo-labels are combined to train the final segmentation model.

Though not explicitly studied in any literature, transfer learning is another approach in this category. In transfer learning, a model is pretrained on a separate set of data and fine-tuned with the images from the target domain. The idea here is that because the model have learned the basics of crack segmentation in the pretraining stage, the number of target data can be reduced.

One major issue with semi-supervised and transfer learning approaches is that for the crack segmentation problem, pixel-precise annotations take disproportionately longer time than weak annotations.
As summarized in \ref{tbl:dataset_overview}, the approximate annotation strategy can reduce the annotation time by 96\%. This means that given the same time budget, semi-supervised and transfer learning approaches can only have access to 4\% of annotated training data. However, the semi-supervised methods mentioned earlier are only evaluated with 12.5\% of annotated data, with a noticeable drop in performance.

\begin{figure}[!tb]
	\centering
	\includegraphics[width=1\linewidth]{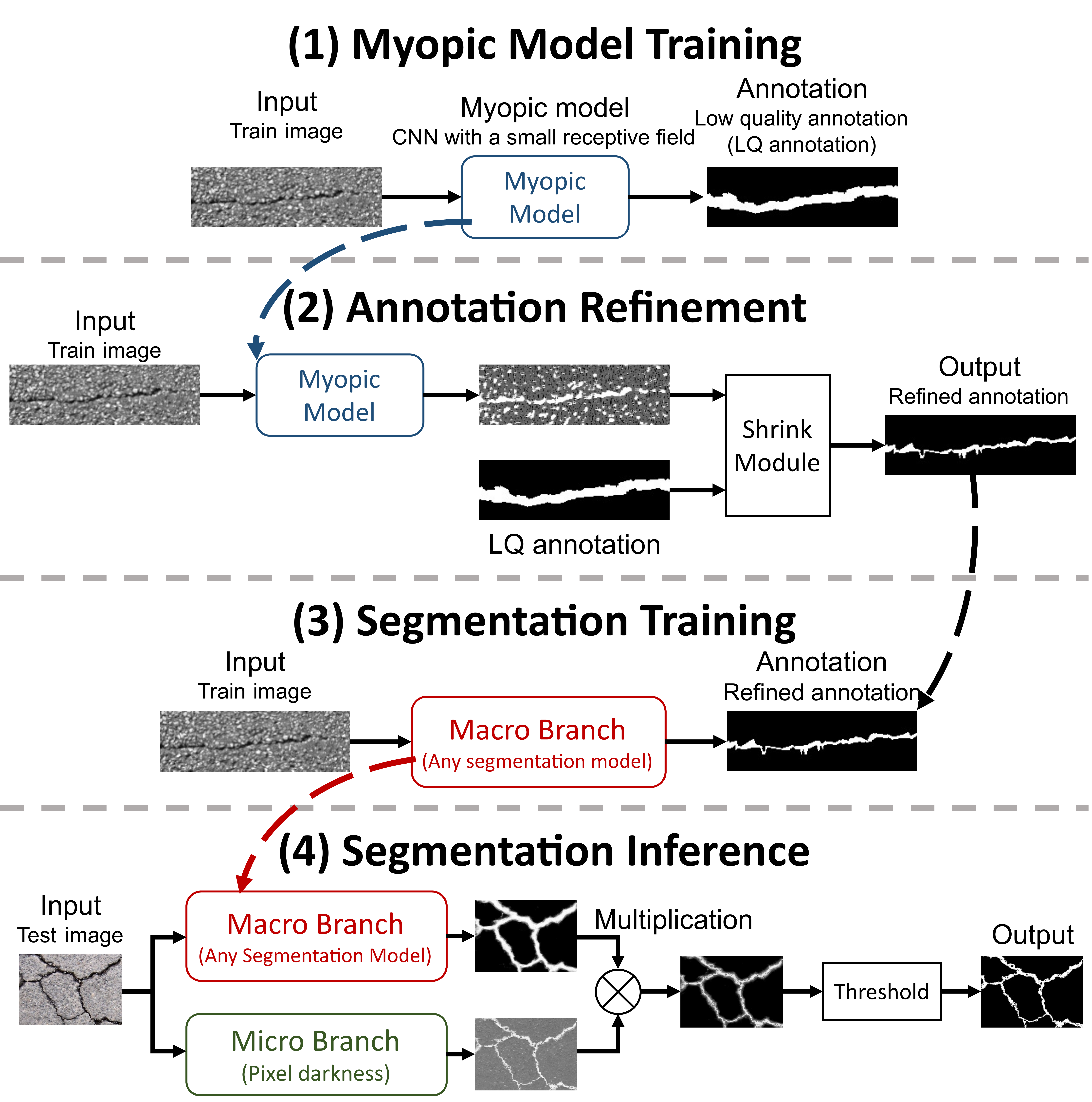} 
	\caption{Overview of the proposed framework.}
	\label{fig:overall}
\end{figure}

\section{Proposed Method}

Formulating crack detection as a pixel-level weakly-supervised problem implies that the annotation labels contain mistakes, mostly concentrated near the crack boundaries. Therefore, the boundary information lost during the annotation process must somehow be recovered to improve the segmentation performance. For this purpose, we propose a multi-step training framework as outlined in \ref{fig:overall}. In this section, we describe two major modifications, annotation refinement and pixel darkness assisted inference, that drive the proposed framework.

\subsection{Annotation Refinement} \label{ssec:refinement}

The first two steps in \ref{fig:overall} are related to refining the low quality (LQ) annotations.
Note that annotation refinement in the context of weakly-supervised crack segmentation is not new- both K\"{o}nig \etal \cite{konig2021weakly} and Dong \etal \cite{dong2020patch} use classification models to generate better annotations.
However, because their methods are based on classification models, they lend themselves better to patch-level annotations. For example, one of the first steps in their method is to subdivide the input images into smaller patches for classifier training. This is done by sampling the annotation with a stride, which is lossless in terms of information for patch-level annotations, but the same procedure loses the free-form information from the pixel-level annotations, which we stated is the advantage of the pixel-level annotations. Also, we later empirically discover in the evaluation that CAM-based methods are bad at utilizing finer annotations.
Therefore, we propose a more suitable annotation refinement strategy for our problem setting consisting of two parts- Myopic Models and Shrink Module, which we will explain in the following paragraphs.

% A major shortcoming of the two-branch inference strategy is that as the annotation quality degrades, the ability of the Macro Branch to locate the cracks degrades proportionally, and it becomes too challenging for the Micro Branch to accurately recover the crack details. Since the rule-based Micro Branch heavily depends on the fact that cracks are dark, this problem is more prominent for datasets that contain cracks of lighter colors.
% One way to remedy this situation is to regain the Macro Branch's ability to locate cracks by somehow improving the annotation quality.
% In this paper, we introduce an annotation refinement process that requires no further additional information, as summarized in \ref{fig:overall} in green.

\begin{figure}[!tb]
    \captionsetup[subfloat]{margin=2pt,format=hang,singlelinecheck=false,font=footnotesize}
    \centering

	\subfloat[Precise annotation.\\ B: true crack, \\ W: true non-crack.]{
	    \includegraphics[width=0.32\linewidth,trim={-2.5cm 0 -2.5cm 0},clip]{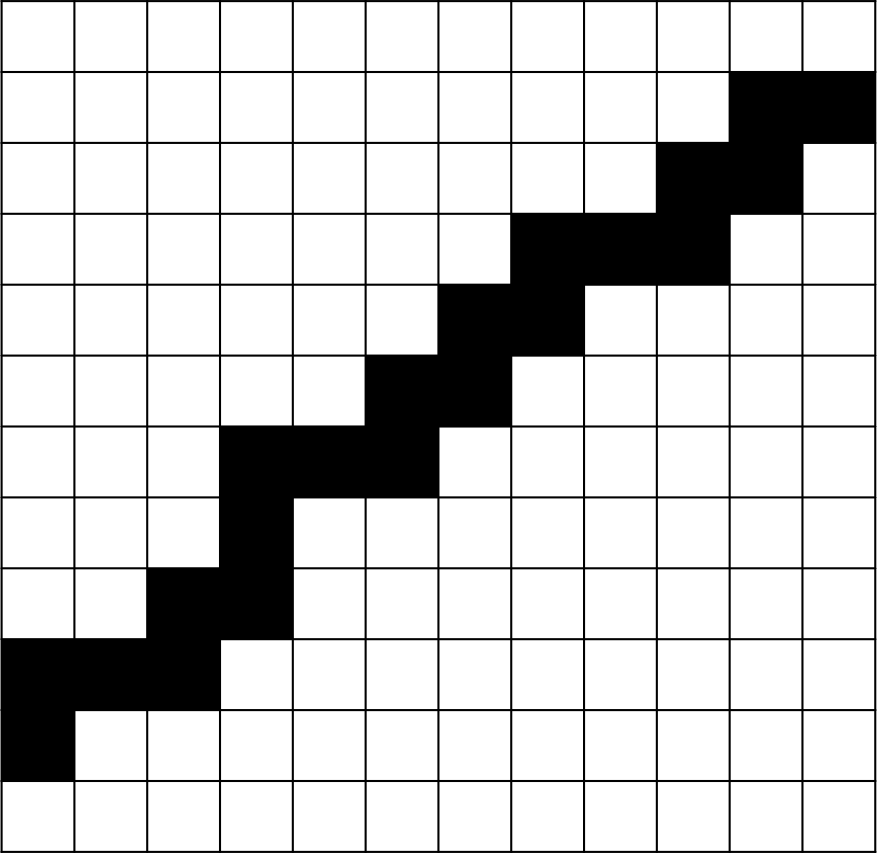}
	    \label{fig:myopia_idea_a}
	}
	\subfloat[LQ annotation. \\ B: LQ crack, \\ W: LQ non-crack.]{
	    \includegraphics[width=0.32\linewidth,trim={-2.5cm 0 -2.5cm 0},clip]{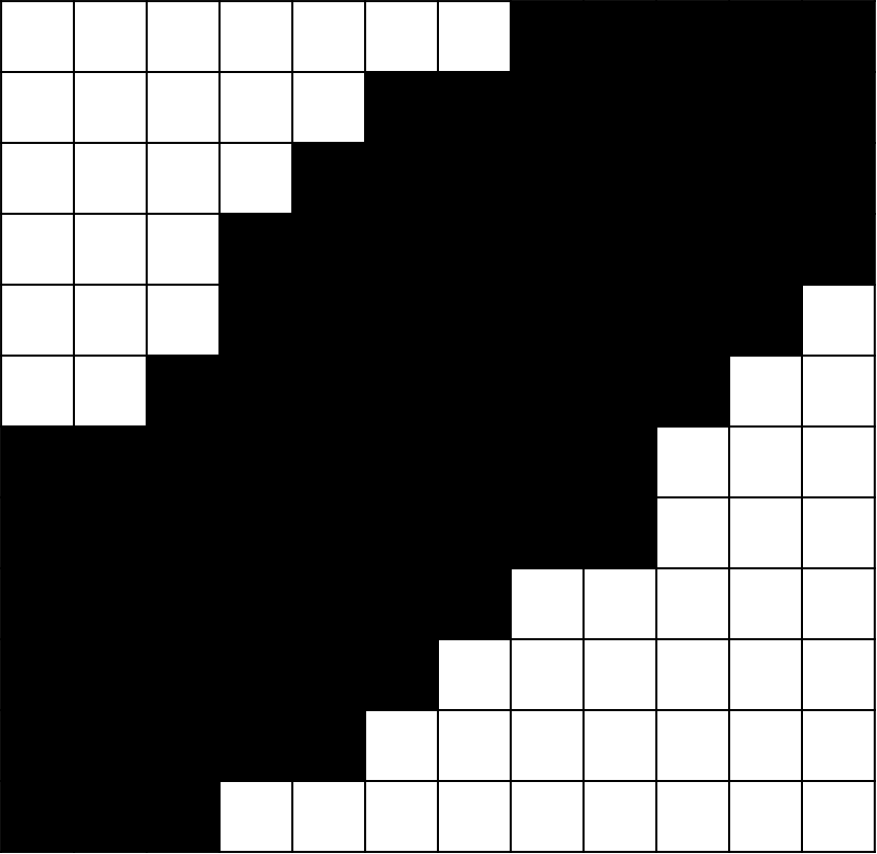}
	    \label{fig:myopia_idea_b}
	}
	\subfloat[Superposed.\\ G: proximity pixels\\(false-positives).]{
	    \includegraphics[width=0.32\linewidth,trim={-2.5cm 0 -2.5cm 0},clip]{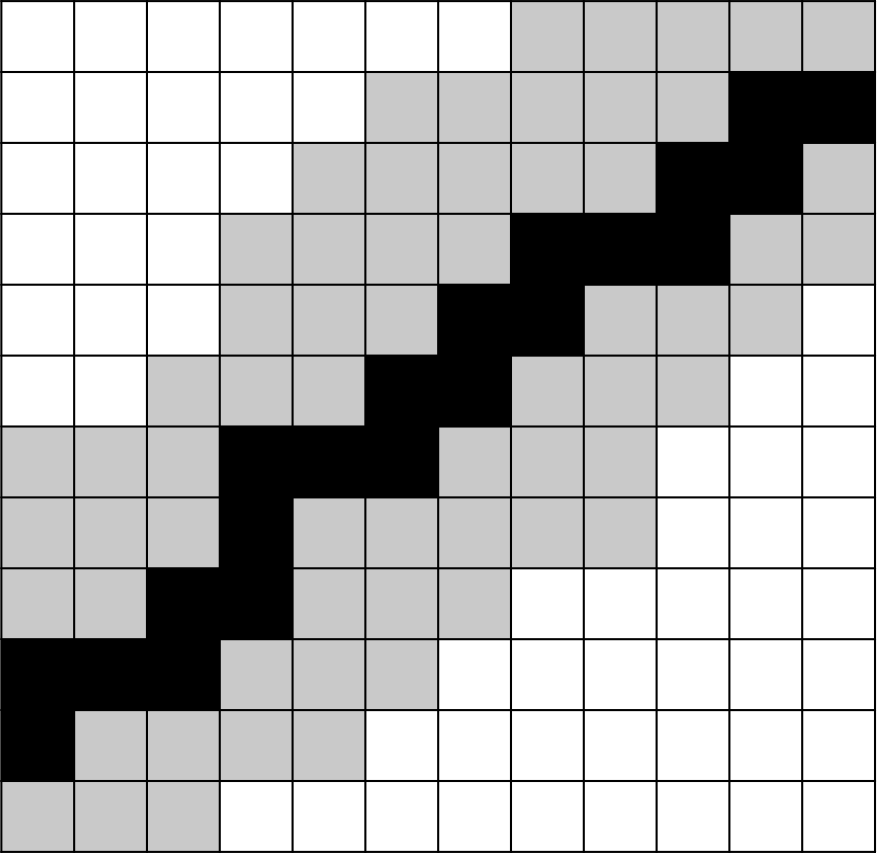}
	    \label{fig:myopia_idea_c}
	}

	\caption{Illustrations of weakly-supervised annotation. Each cell represents a pixel. \textit{LQ} in the captions stand for low quality, and \textit{B}, \textit{W}, \textit{G} stand for black, white, and gray pixels, respectively.}
	\label{fig:myopia_idea}
\end{figure}

\begin{figure}[!tb]
	\centering
	\includegraphics[width=0.8\linewidth]{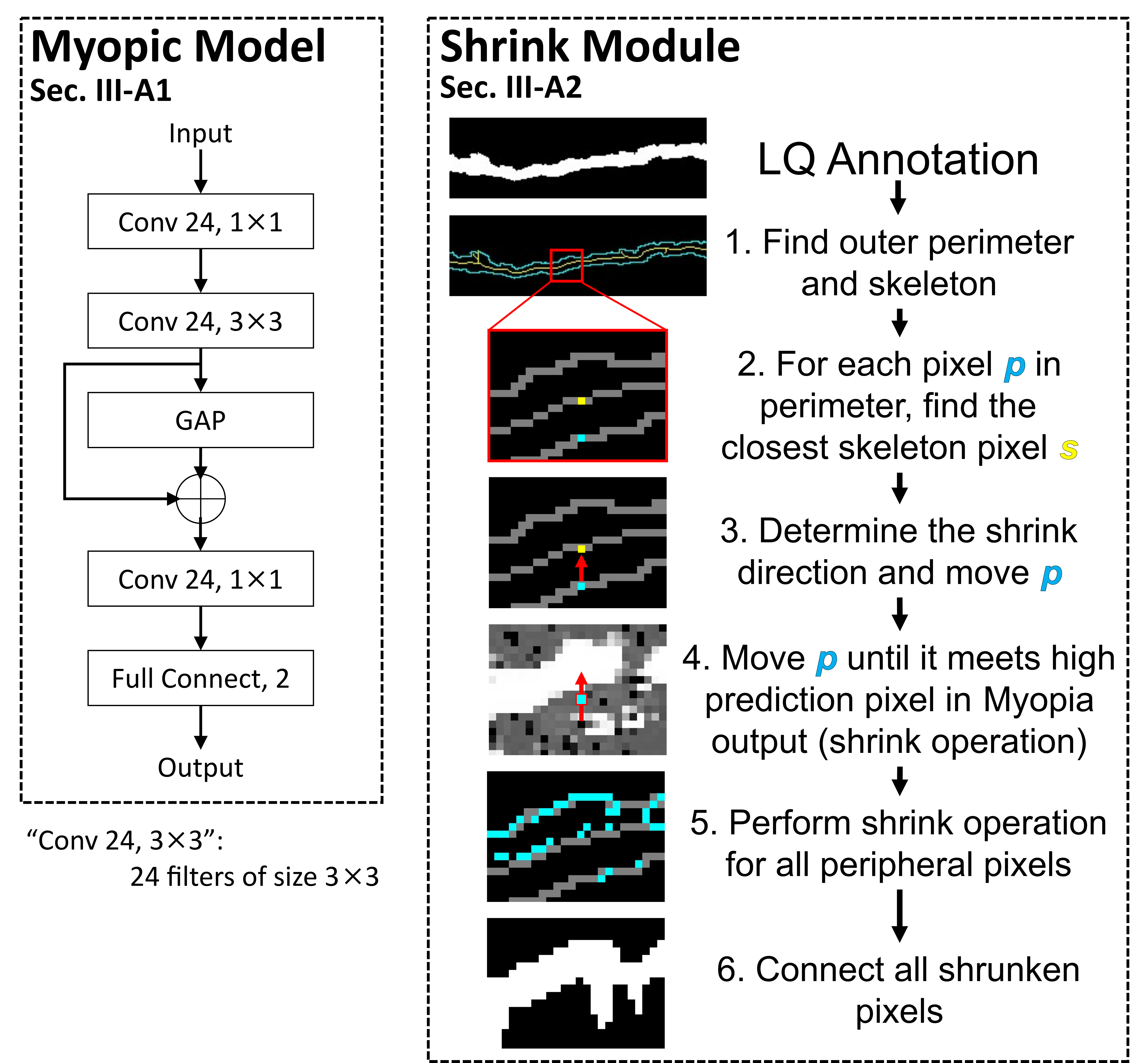} 
	\caption{Two key components of the annotation refinement.}
	\label{fig:overall_sub}
\end{figure}

\subsubsection{Myopic Models} \label{sec:myopic_model}

At first glance, we are tasked with an impossible problem: how can the quality of the LQ annotation be improved, if no extra information is available? To analyze this problem, let us investigate the LQ annotations in closer details. To avoid confusions, the crack and non-crack regions of the precise annotation will be referred to as \textit{true crack} and \textit{true non-crack} regions, and that of the LQ annotation as \textit{LQ crack} and \textit{LQ non-crack} regions.

% This short-sighted nature of the Myopic Models allows them to ignore false-positive regions. The basic idea is illustrated in \ref{myopia_idea}. The blue squares represent the pixel to be predicted, and the red squares represent the corresponding receptive field of the Myopic Models with different $R$.

% For Myopic Models with large $R$ such as \ref{myopia_idea} (b), the receptive field contains the 

% Because the orange square contains true crack pixels in black, the model can see that this region is in proximity to the true crack region. As a result, it is clear to the model that this is an LQ crack region. On the other hand, because the red square does not contain true crack pixels, the model does not know that this region is in proximity to the true crack region. As a result, to the "eyes" of the model, it becomes difficult to make a distinction between the LQ crack region and the non-crack region.

\ref{fig:myopia_idea_a} and \ref{fig:myopia_idea_b} illustrate the examples of fully-supervised and weakly-supervised annotations of the same crack, respectively, where each cell in the grid represents one pixel.
As illustrated in the figure, LQ crack pixels are composed of two types of pixels: those that are truly cracks and false-positive pixels that surround them (\textit{proximity pixels}). Note that we do not consider false-negative pixels because annotators are instructed to over-annotate.
In order to improve the quality of the LQ annotation, a model must classify true crack pixels as cracks and proximity pixels as non-cracks, despite the fact that both are annotated as cracks in the LQ annotation. However, as models typically have large enough receptive fields to judge if a pixel is in the proximity of true crack pixels, they learn to classify proximity pixels as cracks, which is undesired.
The problem here is that a model has too large of a receptive field, so we instead design a segmentation model with a small receptive field, so as to restrict it from being able to judge if a pixel is in the proximity of a crack. Because of this short-sighted nature, we name this model the \textit{Myopic Model}.

Because Myopic Models do not have the access to the proximity information, they cannot differentiate between proximity pixels and LQ non-crack pixels, as the two regions have similar local appearances (since they both do not contain cracks). This results in conflicting backpropagation updates generated by two similarly appearing inputs. However, because there are more LQ non-crack pixels than proximity pixels, the backpropagation signals from the LQ non-crack pixels dominate. As a result, the Myopic Models correctly learn to classify proximity pixels as non-cracks, contrary to what is dictated by the LQ annotation.

We implement the Myopic Model as a simple three-layer CNN as shown left of \ref{fig:overall_sub}, with a receptive field of size $3\times 3$.
Note that as long as the receptive field size is kept small, other forms of implementations such as support vector machines and decision tree-based methods may be just as effective. CNN is chosen merely due to its simplicity in implementation.

% As a result, false-positive regions and LQ non-crack regions should appear similar to Myopic Models.

% This is illustrated in \ref{myopia_idea}. The blue squares represent the pixel to be predicted, and the red squares represent the corresponding receptive field of the Myopic Models with $R=3,9$. Because the true crack pixel is not contained in the receptive field in \ref{myopia_idea}(a), it does not know that the pixel in the blue square is close to true crack regions, making it difficult to tell if it is in LQ non-crack region or false-positive region. On the other hand, because the receptive field contains true crack pixels in \ref{myopia_idea}(b), the model knows that the pixel is close to the true crack region and judges that it is not an LQ non-crack pixel.

% There is one drawback with Myopic Models, however. The small receptive field is a blessing and a curse, as it trades off refinement power for detection accuracy. Empirically we found that the Myopic Models' raw predictions are full of false-positives, but these new false-positives tend to exist relatively far away from true crack regions, so they can be masked away by the original low-quality data.

To further strengthen the ability of the Myopic Model to ignore mislabels during training, the cross-entropy loss is modified as shown in \ref{L_CE}. In the equation, $x$ is a pixel in an image, $p_x$ is the predicted crack probability at $x$, $H$ represents the set of pixels in an LQ crack region with crack probability in the top 90 percentile, and $B$ represents the set of pixels in an LQ non-crack region with crack probability in the bottom 90 percentile. The added conditions (in colors) will be referred to as the ignore conditions, as they force the model to ignore its low predictions in the crack regions (red) and high predictions in the non-crack regions (blue).
% The ignore conditions account for the fact that low-quality annotations contain annotation errors, and prevent the model from receiving incorrect supervisions.

\begin{equation}
    L_{CE} = -
    \tcboxmath[
        colorbox,
        % overlay={
        %     \node[below,red,text width=4cm,align=center] at (frame.south) {Crack pixels with top 90\% probability output};
        % }
    ]
    {\sum_{\substack{x\in H}}
    } \log(p_x)
    -
    \tcboxmath[
        colorbox=blue!10,
        % overlay={
        %     \node[below,red,text width=4cm,align=center] at (frame.south) {Crack pixels with top 90\% probability output};
        % }
    ]
    {\sum_{\substack{x\in B}}}
    \log(1-p_x)
    \label{L_CE}
\end{equation}

\subsubsection{Shrink Module} \label{subsec_anno_refine}

We discussed in previous paragraphs that the mislabel-ignoring nature of the Myopic Models can be used to refine the LQ annotations.
However, we observed that a simple application of the Myopic Model not only removes false-positives from the annotation but also true crack pixels as well, introducing false-negatives. The false-negatives typically break continuities of crack regions, which is undesired for training. In order to preserve continuity, we introduce the Shrink Module.

Assuming that the LQ crack regions completely cover the true crack regions, annotations can be refined by \textit{shrinking} the contours of the LQ crack regions. As outputs of a Myopic Model provide good indications of where true crack pixels are, they are used to guide the shrinking process. In addition, because this shrink procedure preserves cracks' continuity, it should produce better refinement results than simply using the Myopic Model's output as the refined annotation.

The right side of \ref{fig:overall_sub} summarizes the details of the Shrink Module.
The process starts by determining which pixels in the LQ crack annotation $L$ the shrink operation should be applied to, by calculating the outer contour pixels\footnote{Calculated via \textit{findContours} function in OpenCV package \cite{opencv_library}.} $P$ (step 1).
Then, the shrink direction $v$ is calculated from the closest point between $p \in P$ and skeletonized pixel\footnote{Calculated via \textit{skeletonize} function in scikit-image package \cite{scikit-image}.} $s \in S$ (steps 2 and 3).
Then, each $p$ is moved in $v$ direction until it meets a pixel with a high probability in the Myopic Model output $M$ (step 4, \textit{shrink operation}). ``High probability'' is defined as the case in which the probability value of $M$ at the new pixel is higher than that at the original contour coordinate by a threshold.
To reduce noise, the shrink operation continues until the condition is met twice in a row.

The shrink operation is performed for all $p$ to form the refined set $R$ (step 5). If the criterion is not met for a particular $p$, its starting coordinate is added to $R$.
The final refined image is generated by connecting the pixels in $R$ and filling the resulting contour.

The output of the Shrink Module is the final refined annotation, used in step 3 in \ref{fig:overall} to train the \textit{Macro Branch}, which can be any off-the-shelf segmentation model.

% Finally, the step-by-step procedure of the proposed framework is summarized below:

% \begin{enumerate}
%     \item Train a Myopic Model $f_{mm}$, using LQ annotation $D_{lq}$ (\ref{sec:myopic_model})
%     \item Generate refined annotations $D_{ref}$ from the outputs of $f_{mm}$ and $D_{lq}$, via Shrink Module (\ref{subsec_anno_refine})
%     \item Train a Macro Model $f_{macro}$ using $D_{ref}$
%     \item Perform inference using $f_{macro}$ and $f_{micro}$ (rule-based darkness calculation, \ref{ssec:darkness})
% \end{enumerate}

\subsection{Pixel Darkness Assisted Inference} \label{ssec:darkness}

We further improve the segmentation result of the Macro Branch by adding another inference branch named the \textit{Micro Branch}, which simply outputs per-pixel darkness value as the crack probability (step 4 in \ref{fig:overall}).
This is based on the observation that human annotators annotate cracks in two steps. First, an annotator determines the rough locations of cracks by examining the entire image. Then, the annotator zooms into a section, compares its pixel darkness against its neighbors, and annotates dark pixels as cracks.
The roles of the two branches emulate these steps- the Macro Branch determines the rough crack locations, and the Micro Branch recovers the fine details of cracks.
The outputs of two branches are aggregated by a pixel-level multiplication and thresholded to produce the final output.

\subsection{Pixel Darkness vs. Annotation Refinement} \label{ssec:mib_vs_refine}

We conclude this section by comparing the annotation refinement process (\ref{ssec:refinement}) and the pixel darkness approach (\ref{ssec:darkness}).

In both methods, certain inductive biases about the segmentation target are made in order to recover the target boundary information lost during the imprecise annotation process.
The pixel darkness based approach assumes that the targets are darker, which is highly effective for finding cracks, but also inflexible as the decision logic is fixed regardless of the target.
On the other hand, the annotation refinement approach makes three assumptions about the target: (1) the background region is significantly larger than the target region (2) the target has a small chromatic variation (3) the target is long and connected. The first two assumptions are utilized in designing the Myopic Model and the last assumption is utilized in designing the Shrink Module. These assumptions are more general than the pixel darkness assumption, as it is valid for any thin targets, not just for cracks. Furthermore, the chromatic properties of the target are learned from the dataset by the Myopic Model, making it more robust to different targets.

These assumptions also illustrate the limitations of the proposed method.
First, the Micro Branch fails when the target is not dark. Second, the annotation refinement pipeline fails when the target object is not thin or monochromatic.

\section{Experiments}

\begin{figure*}[!tb]
    \small
    \centering
    \renewcommand{\arraystretch}{0.6}
    \setlength{\tabcolsep}{1pt}
    \newcolumntype{C}{>{\centering\arraybackslash} m{2.3cm} }
    \newcolumntype{B}{>{\centering\arraybackslash} m{0.4cm} }
	\begin{tabular}{BCCCCCC}

    \multirow{8}{*}{\rotatebox[origin=c]{90}{Aigle}} &
    \includegraphics[width=\linewidth]{Im_GT_AIGLE_RN_F14aor_output/detailed_original.png} &
    \includegraphics[width=\linewidth]{Im_GT_AIGLE_RN_F14aor_output/detailed_gt.png} &
    \includegraphics[width=\linewidth]{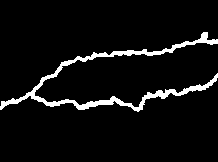} &
    \includegraphics[width=\linewidth]{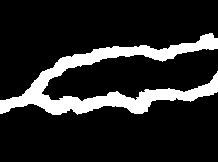} &
    \includegraphics[width=\linewidth]{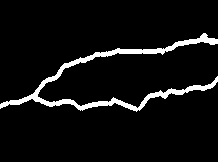} &
    \includegraphics[width=\linewidth]{Im_GT_AIGLE_RN_F14aor_output/rougher_gt.png} \\

    &
    \includegraphics[width=\linewidth]{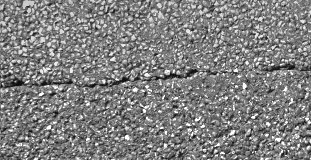} &
    \includegraphics[width=\linewidth]{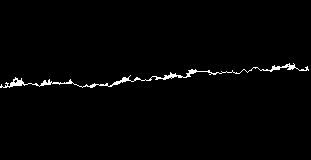} &
    \includegraphics[width=\linewidth]{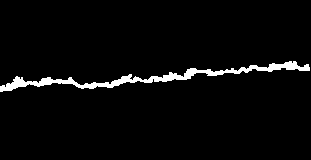} &
    \includegraphics[width=\linewidth]{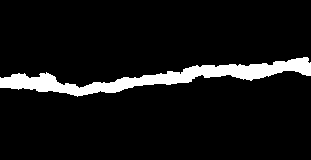} &
    \includegraphics[width=\linewidth]{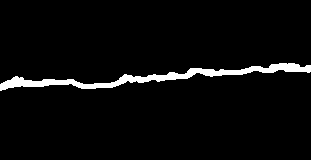} &
    \includegraphics[width=\linewidth]{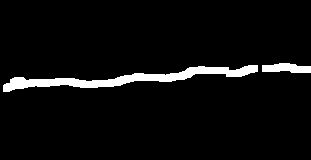} \\

    \multirow{8}{*}{\rotatebox[origin=c]{90}{CFD}} &
    \includegraphics[width=\linewidth]{036_output/detailed_original.png} &
    \includegraphics[width=\linewidth]{036_output/detailed_gt.png} &
    \includegraphics[width=\linewidth]{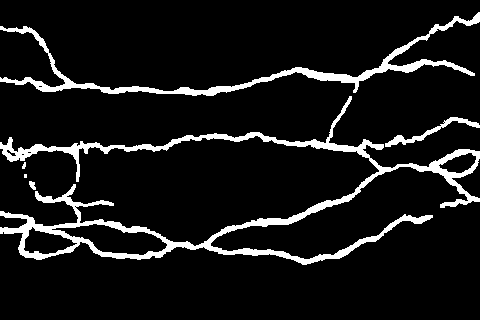} &
    \includegraphics[width=\linewidth]{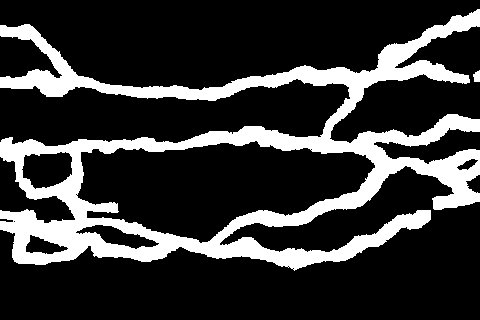} &
    \includegraphics[width=\linewidth]{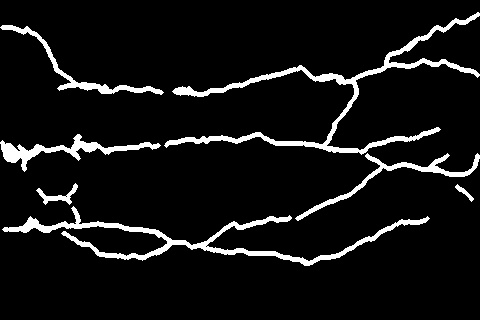} &
    \includegraphics[width=\linewidth]{036_output/rougher_gt.png} \\

    &
    \includegraphics[width=\linewidth]{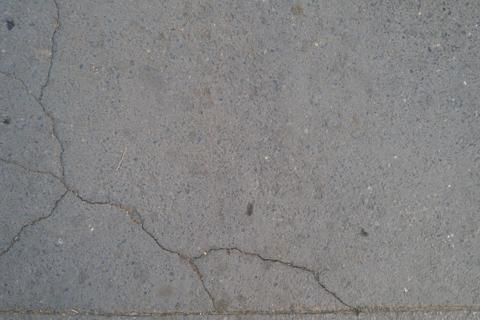} &
    \includegraphics[width=\linewidth]{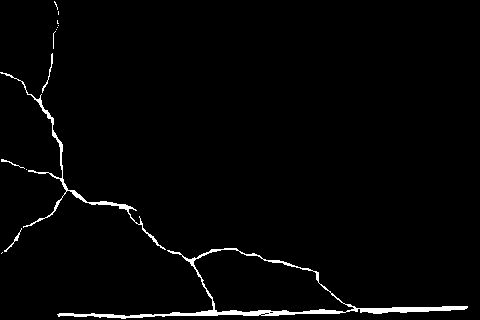} &
    \includegraphics[width=\linewidth]{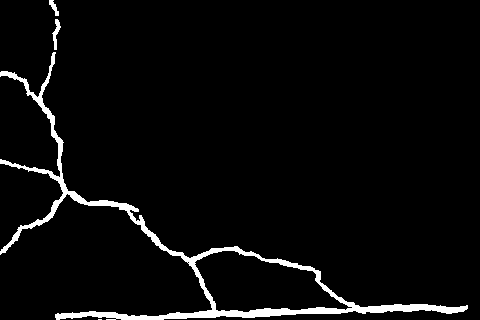} &
    \includegraphics[width=\linewidth]{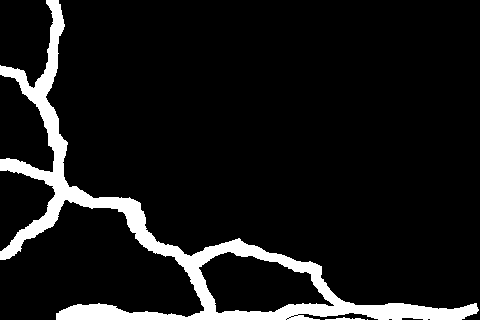} &
    \includegraphics[width=\linewidth]{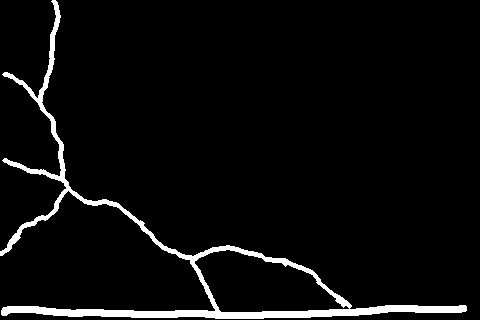} &
    \includegraphics[width=\linewidth]{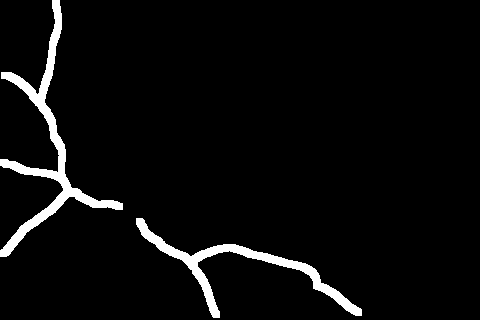} \\

    \multirow{8}{*}{\rotatebox[origin=c]{90}{DCD}} &
    \includegraphics[width=\linewidth]{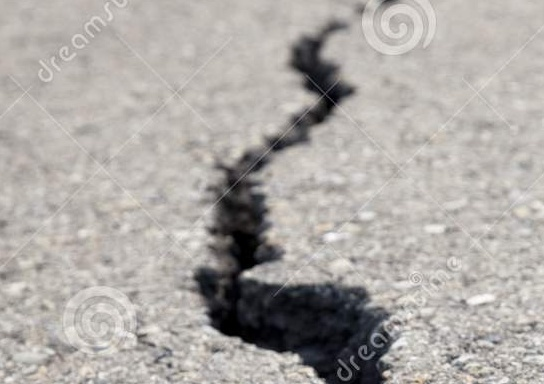} &
    \includegraphics[width=\linewidth]{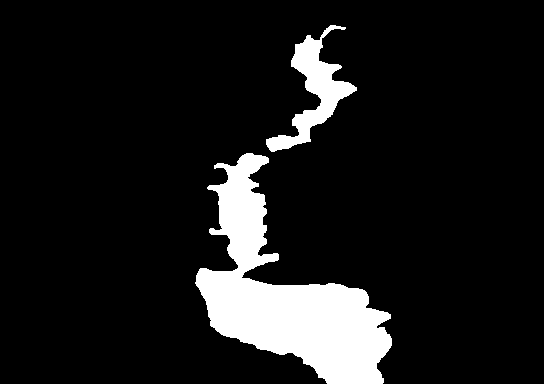} &
    \includegraphics[width=\linewidth]{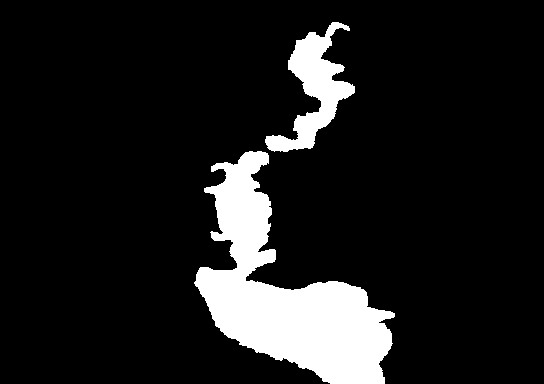} &
    \includegraphics[width=\linewidth]{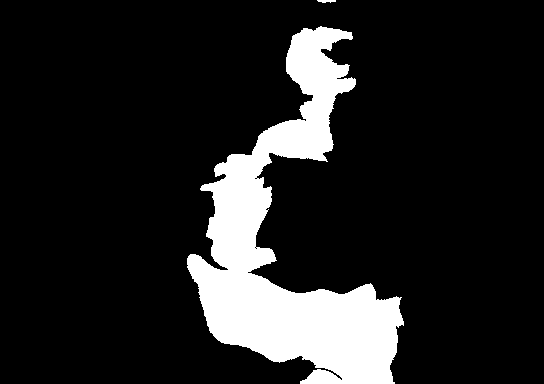} &
    \includegraphics[width=\linewidth]{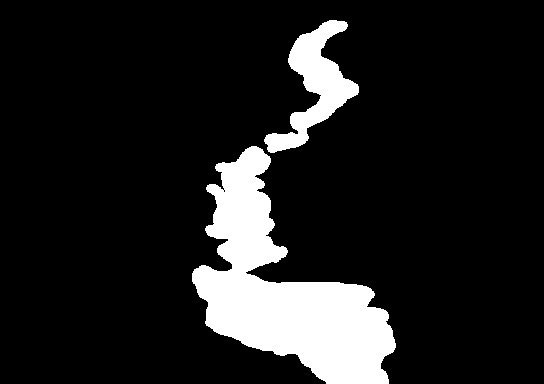} &
    \includegraphics[width=\linewidth]{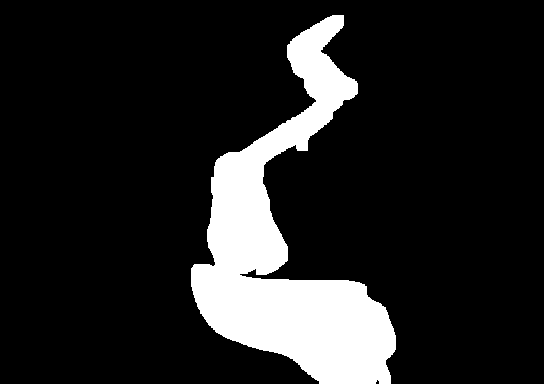} \\

    &
    \includegraphics[width=\linewidth]{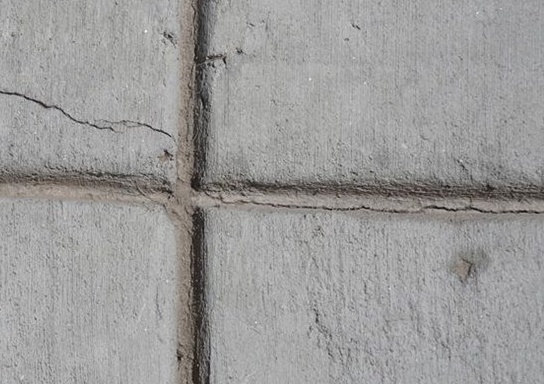} &
    \includegraphics[width=\linewidth]{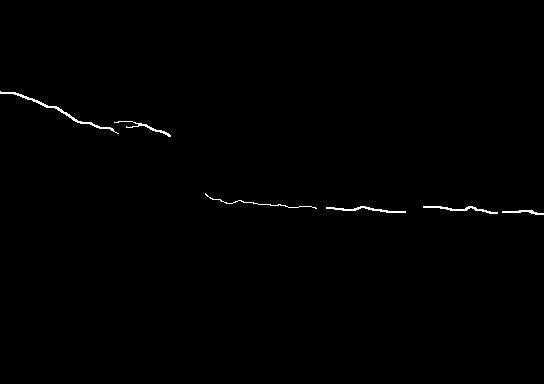} &
    \includegraphics[width=\linewidth]{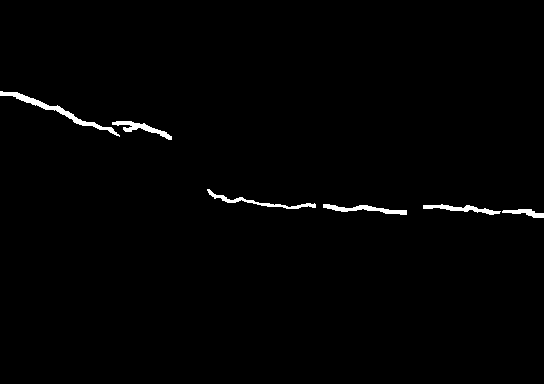} &
    \includegraphics[width=\linewidth]{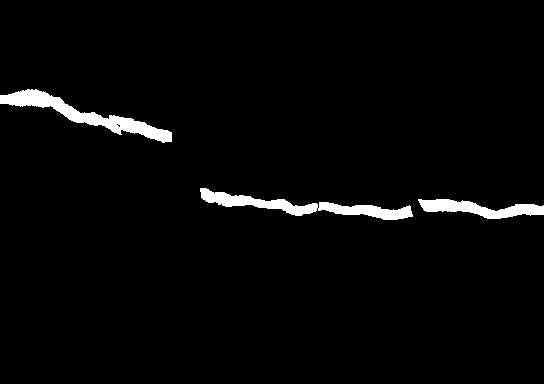} &
    \includegraphics[width=\linewidth]{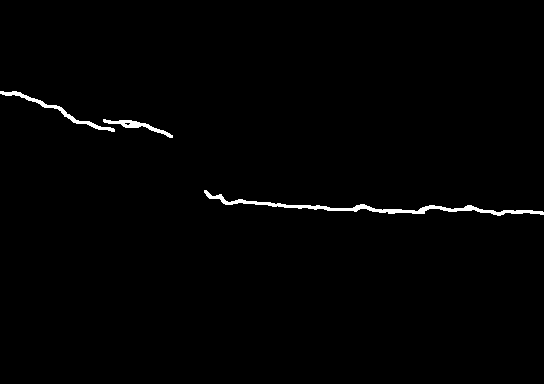} &
    \includegraphics[width=\linewidth]{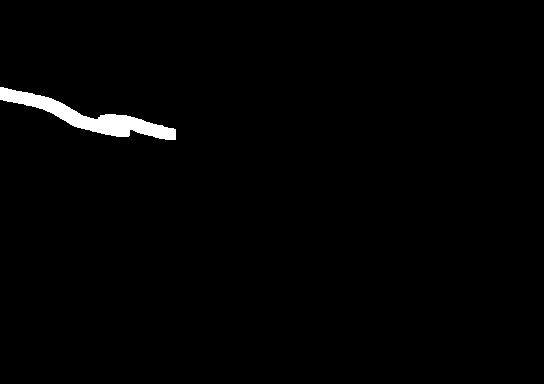} \\

    \\[-1.5mm]
     & Original & Precise & Dil-1 & Dil-4 & Rough & Rougher \\

    \end{tabular}
    \caption{Annotation samples (white pixels: crack, black pixels: non-crack). Taken from \cite{inoue2020crack}. Dil-1 and Dil-4 correspond to synthetic annotations with 1 and 4 dilations during the synthesis process, respectively.}
    \label{fig:gt_samples}
\end{figure*}

\subsection{Dataset} \label{ssec:dataset}

The Aigle dataset \cite{aiglern}, Crack Forest dataset (CFD) \cite{cfd}, and DeepCrack dataset (DCD) \cite{liu2019deepcrack} are used for evaluation. As shown in \ref{fig:gt_samples}, each dataset has distinct characteristics.
\subsubsection{Aigle} It is the smallest dataset, selected to evaluate the proposed method under a data-scarce situation. Also, as the figure shows, Aigle consists of asphalt surfaces, which have a complex and noisy background that could be mistaken as cracks.
\subsubsection{CFD} It is the most commonly studied dataset in crack segmentation according to \cite{huang2022nha12d}, and it is also the reason why it was selected for evaluation. CFD consists of road-surface images taken in Beijing using a smartphone camera.
\subsubsection{DCD} It consists of 537 images, making it one of the larger crack datasets. A unique characteristic of DCD is that it contains cracks formed in various surfaces, including everything from walls, pavement to tiled floors. Compared to most other crack datasets which are primarily composed of road cracks, DCD provides an opportunity to conduct evaluations under a variety of environments.

\subsection{Annotation Preparation} \label{ssec:annotation}

In addition to the pixel-accurate annotations provided with the datasets (Precise Annotation), two types of imprecise annotations, manual and synthetic, are prepared.

\subsubsection{Manual Annotation} \label{ssec:manual_anno}

Manual annotations are annotated by a human annotator and there are two types: Rough and Rougher, with Rougher lower in quality.
The annotation rules provided to the annotator are as follows (values in \{\} correspond to rules for the Rough Annotation, and [] correspond to rules for the Rougher Annotation):

\begin{quote}
Set the size of the pen tool to be \{1 or 2\} [3 or 4] pixels larger than the average width of the cracks. You are allowed to adjust it \{as many times as needed\} [once at most] per image. Trace the cracks in one stroke unless the crack width is larger than the pen size, and \{follow as much as possible\} [ignore] the small contours.
For thicker cracks (approx. 8 pixels or wider), use a pen size of \{4\} [8] pixels to trace the outline. Fill it with a bucket tool afterward.
\end{quote}

The time taken for annotation are summarized in \ref{tbl:dataset_overview}. Note that the annotation times for the Precise Annotation are unknown, so CFD was re-annotated precisely for approximation. As the table shows, it took an order of magnitude shorter to annotate the Rough and Rougher Annotations than to annotate the Precise Annotation.

\subsubsection{Synthetic Annotation} \label{ssec:syn_anno}

In addition to manual annotations, synthetic annotations are machine-generated from Precise Annotation using image dilation and deformation. The synthesized annotations are prepared for three reasons. First, the annotation quality is easily quantifiable and controllable by the number of dilations. Second, the synthesis pipeline can arbitrarily be extended to obtain annotations of various qualities at a small cost. Finally, it is void of any biases that a human annotator may introduce.

Here, we roughly outline the synthesis process.
First, a Precise Annotation sample $p$ is dilated $n_{dil}$ times to generate a dilated sample $d$. This $n_{dil}$ value dictates the quality of the resulting annotation. To test the proposed framework under various settings, four annotations with different $n_{dil}$ are synthesized ($n_{dil} \in \{1,2,3,4\}$).
Then, Elastic Transform \cite{simard2003best} is applied to the dilated sample to form a synthesis candidate $s$. This image deformation step ensures that the true crack region does not always lie in the center of the synthesized weak annotation. Elastic Transform implementation in Albumentations \cite{buslaev2020albumentations} has three parameters: $\alpha$, $\sigma$, alpha\_affine. $\sigma$ is fixed at 12 to prevent excessive deformation, and alpha\_affine is fixed at 0.2.
Finally, $s$ and $p$ are compared by calculating the recall value $r$. In order to emulate rushed human annotators, $r$ should neither be too high nor too low. For the annotations generated for the paper, the upper and the lower bounds are chosen to be 0.975 and 0.925, respectively.
The relationship between values of $\alpha$ and $r$ varies greatly across different images, so $\alpha$ is searched randomly from a uniform distribution, with the range initialized to be between 10 to 10000, and narrowed according to the obtained $r$ value.

\begin{figure*}[!tb]
    \captionsetup[subfloat]{margin=10pt,format=hang,singlelinecheck=false,font=footnotesize}
    \centering
    
	\subfloat[Results on the synthetic annotation. Horizontal axis: annotation quality, lower the value, higher the quality. \textit{0} corresponds to the Precise Annotation.]{
	    \includegraphics[width=0.45\linewidth]{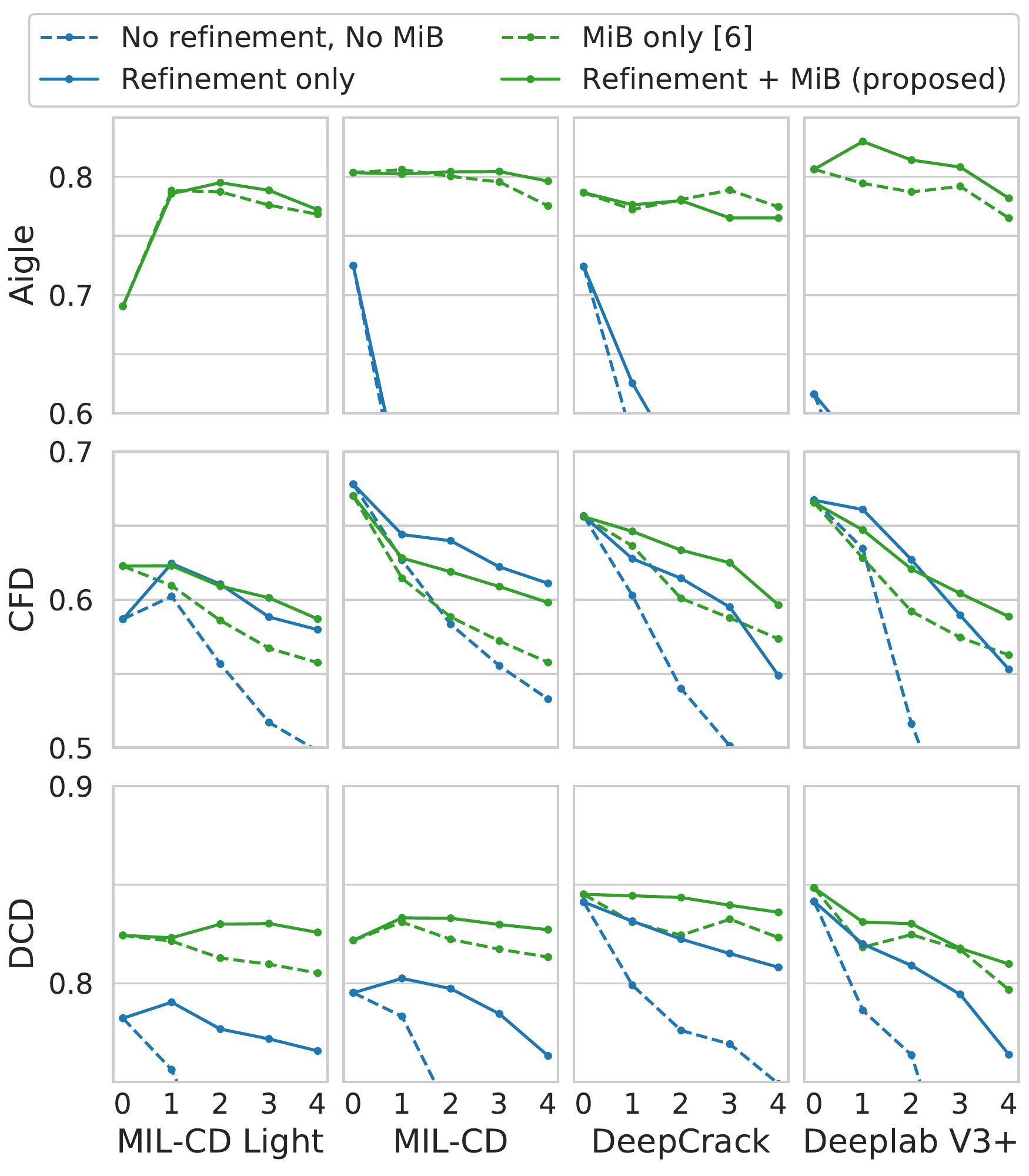}
        \label{fig:eval_results_syn}
	}
% 	\hspace{3mm}
	\subfloat[Results on the manual annotation. P: Precise, R: Rough, and R-er: Rougher Annotations.]{
	    \includegraphics[width=0.45\linewidth]{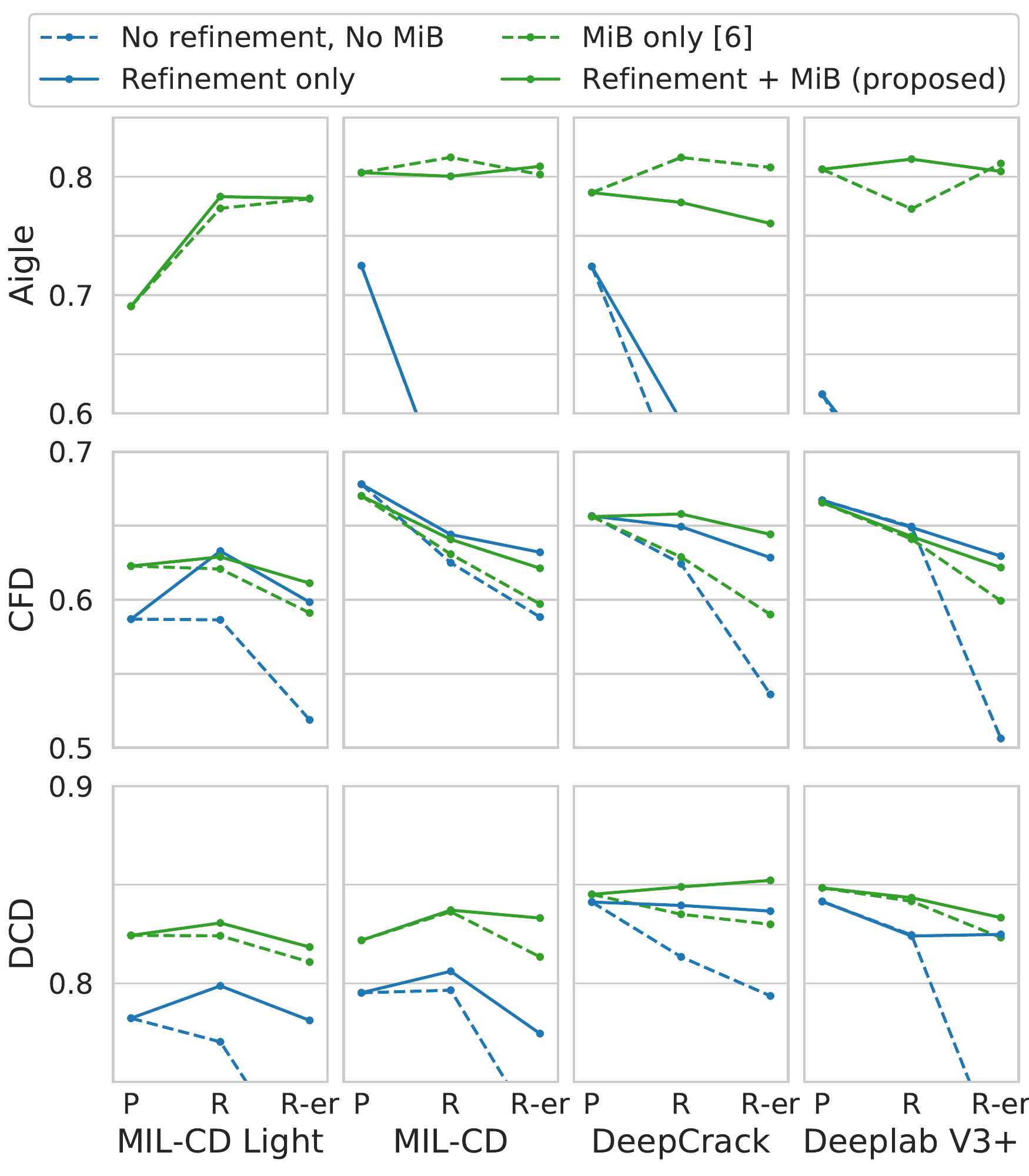}
        \label{fig:eval_results_man}
	}

    \caption{Evaluation results on different LQ annotations. For all plots, annotation quality decreases from left to right. MiB represents cases in which the Micro Branch is included during inference.}
    \label{fig:eval_results}
\end{figure*}

\subsection{Evaluation Metrics}

The models are evaluated using the Optimal Dataset Scale (ODS), as defined in \ref{eq:ods}, where $P^t_i$ and $R^t_i$ represent the precision and recall values at threshold $t$, of the $i$-th image in the dataset of $N$ samples. Since what follows the maximum operator is the definition of the F1-score, ODS is essentially F1-score optimized for the best threshold value to be used in a dataset.

\begin{equation}
    ODS = \max_{0<t<1} \frac{1}{N} \sum_{i=1...N}\frac{2 P^t_i R^t_i}{P^t_i + R^t_i}
    \label{eq:ods}
\end{equation}

\subsection{Macro Branch Implementations}

Four semantic segmentation models are tested as the Macro Branch to assess the versatility of the proposed framework. 

\paragraph{MIL-CD \cite{inoue}}
The model proposed by Inoue \etal achieved high performances for Aigle and CFD. It consists of 7 convolutional layers with around 12 filters each, and it is augmented by the Multiple Instance Learning (MIL) architecture, which calls for the inference to be performed twice, rotating the image by 0$^\circ$ and 90$^\circ$. To avoid confusion, we will refer to this model as MIL Crack Detector (MIL-CD).

\paragraph{MIL-CD Light}
A computationally lighter version of MIL-CD is also tested to evaluate the effectiveness of the framework for extremely light models. This model also has 7 convolutional layers, but \textit{CONV 2} and \textit{CONV 4} layers have strides of 2 instead of 1, the number of filters at each layer is halved, and the MIL configuration is not used.

\paragraph{DeepCrack \cite{liu2019deepcrack}}
The model proposed by Liu \etal achieved high performance for DCD.
It employs encoder-decoder architecture with skip connections, and its inner feature representations are supervised at different stages to better capture information from multiple scales\footnote{Implementation adopted from https://github.com/yhlleo/DeepSegmentor}.

\paragraph{DeepLab V3+ \cite{deeplabv3p}}
This model is one of the state-of-the-art model architectures for the general semantic segmentation task. It employs both spatial pyramid pooling as well as encoder-decoder architecture to capture multi-scale information. Note that this model is by far the most computationally heavy model among the tested models\footnote{Implementation adopted from https://github.com/tensorflow/models/}.

% Experimental parameters of the above models are summarized in \ref{model_settings}.

% First, DeepCrack by Liu \etal \cite{liu2019deepcrack} and the model proposed by Inoue \etal \cite{inoue}, which are both architectures specifically designed for crack detection, were selected to assess the effectiveness of the proposed framework for task-specific models. In addition, DeepLab V3+ \cite{deeplabv3p}, which is designed for general semantic segmentation tasks, was selected to evaluate the proposed framework on general models. Finally, a computationally lighter version of Inoue \etal's model (Inoue Light) was selected to evaluate the framework on computationally lighter models.

% \begin{table}
% 	\centering
% 	\setlength{\tabcolsep}{4pt}
%     \renewcommand{\arraystretch}{1.1}
% 	\caption{Training parameters of the tested models. Inoue Light model shares the same parameters as Inoue \etal.}
% 	\begin{tabular}{lcccl}
% 		\hline
% 		Model       & lr    & \# epochs & $w_c^{model}$ & Notes \\ \hline
% 		Inoue \etal & 1e-1  & 50        & 20  	        & lr halved every 20 epochs \\ \hline
% 		DeepCrack   & 1e-3  & 700       & 33.3          & lr decayed after 400 epochs \\ \hline
% 		\multirow{2}{*}{DeepLab v3+} & \multirow{2}{*}{5e-3}  & \multirow{2}{*}{200000}    & \multirow{2}{*}{100}           & Xception 65 backbone \cite{chollet2017xception}, \\
% 		            &       &           &               & PASCAL VOC pretrained \\ \hline
% 	\end{tabular}
% 	\label{model_settings}
% \end{table}

\begin{table}
	\centering
	\setlength{\tabcolsep}{4pt}
    \renewcommand{\arraystretch}{1.1}
	\caption{Performance improvements for Micro Branch (MiB) and Annotation Refinement compared to full supervision, averaged across all models and weak annotations. Positive values mean better performance than full supervision despite being trained on LQ annotations.}
	\begin{tabular}{lSSS}
	    \toprule
         & {Aigle} & {CFD} & {DCD} \\
        \midrule
        % Fully-Supervised                    &0.636&0.647&0.815 \\
        % \midrule
        No MiB, No Anno. Refine              &-0.188&-0.099&-0.074 \\
        MiB only \cite{inoue2020crack}      &0.153&-0.051&0.006 \\
        Anno. Refine only                    &-0.111&-0.031&-0.015 \\
        MiB + Anno. Refine (Proposed)        &0.156&-0.026&0.018 \\
        \bottomrule
	\end{tabular}
	\label{tbl:improvements}
\end{table}

\begin{figure}[!tb]
    \centering
    \includegraphics[width=0.8\linewidth]{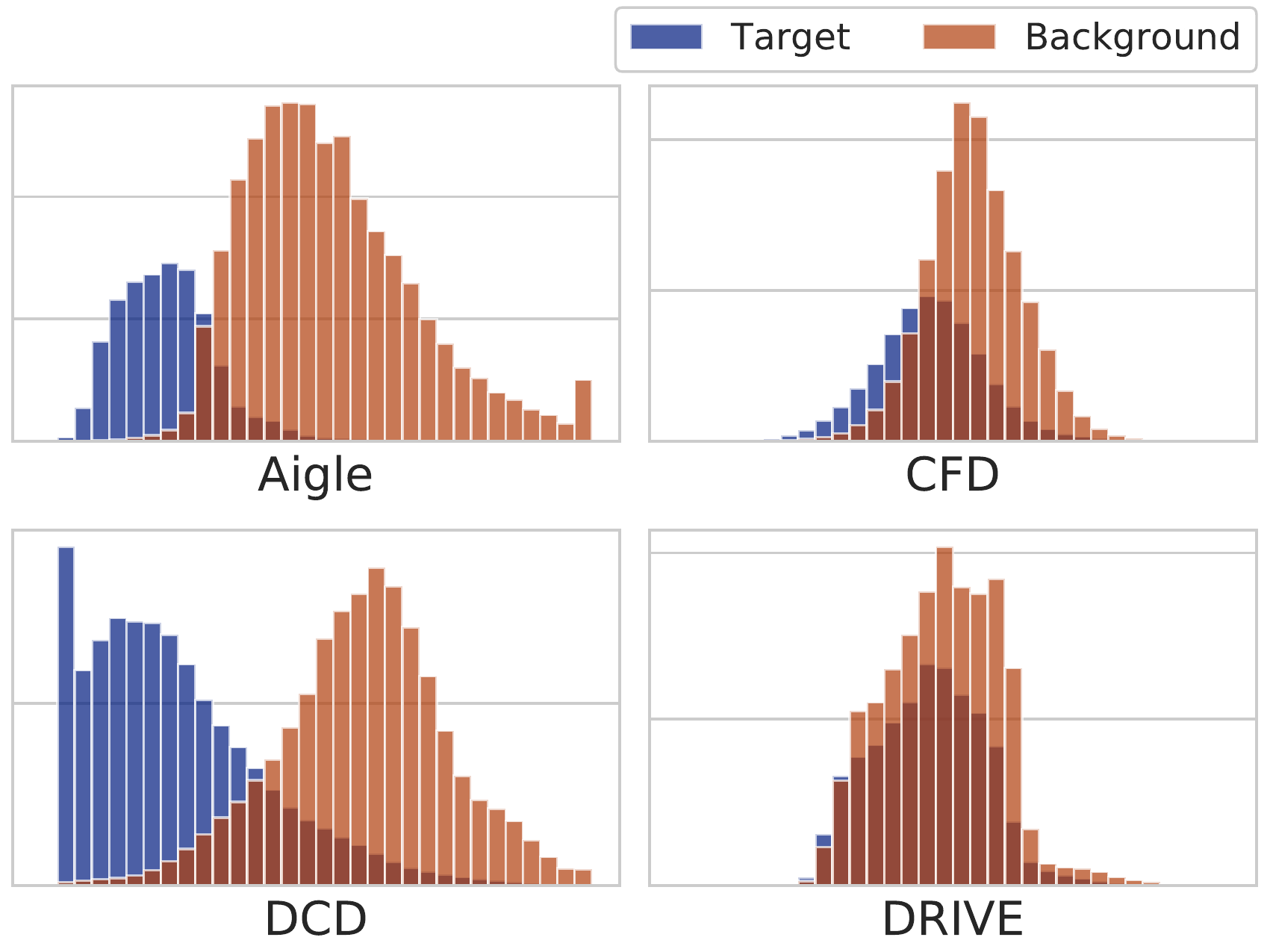}
    \vspace{-3mm}
    \caption{Brightness distribution of the pixels included in the positive regions of Rougher Annotation.}
    \vspace{-2mm}
    \label{fig:histogram}
\end{figure}

\subsection{Evaluation on the Low Quality Annotations} \label{ssec:eval_lq}

The results of the proposed framework on weak supervision are summarized in \ref{fig:eval_results}. The colors of the plots correspond to the presence of the Micro Branch, and the solid and dotted lines correspond to the inclusion of the annotation refinement step.
Note that what we proposed in \cite{inoue2020crack} corresponds to dotted green lines. Furthermore, \ref{fig:eval_results_syn} is the results on the synthetic annotation, with the horizontal axis corresponding to the dilation factor $n_{dil}$ mentioned in \ref{ssec:syn_anno}, and \ref{fig:eval_results_man} is the results on the manual annotation, with the horizontal axis corresponding to Precise (P), Rough (R), and Rougher (R-er) Annotations. Also note that \textit{0} and \textit{P} in \ref{fig:eval_results_syn} and \ref{fig:eval_results_man} respectively correspond to results on fully-supervised setting.
The plots show that annotation refinement improves model performance for most datasets and models, confirming its effectiveness. In addition, annotation refinement is effective regardless of Micro Branch presence, which means that the two methods improve the performance independently.

To further analyze the effect of the Micro Branch and annotation refinement, the ODS performances are averaged across all models and weak annotations (both manual and synthetic) and compared against the fully-supervised performance. Results are summarized in \ref{tbl:improvements}.
Since the values in the table are calculated as (averaged ODS on weak annotations) - (full supervision ODS), negative values are expected. For example, when the Micro Branch nor annotation refinement is applied, the performances drop between 7.4\% to 18.8\%.
The goal of weakly-supervised methods is to improve this value to 0, as the value 0 indicates a full recovery of the model performance despite trained with LQ annotations.

The table shows that for Aigle and DCD, the Micro Branch is enough to fully recover the performance. This is because the pixel brightness distribution is well-separated for those datasets, as shown in \ref{fig:histogram}. Because Micro Branch places a strong assumption on pixel darkness as discussed in \ref{ssec:mib_vs_refine}, it works extremely well for the two datasets. This is why we claimed in \ref{sec:intro} that weakly-supervised problem for this kind of situation is already solved.
However, the Micro Branch does not perform nearly as well for CFD- it is only able to halve the performance drop from 9.9\% to 5.1\%.

The story is different for annotation refinement- it is able to consistently improve model performance by 6-7\%, including on CFD. Although there is still room for improvement, it was able to halve the performance gap between fully-supervised and Micro Branch only methods (\ie $-3.1\%$ as opposed to $-5.1\%$).
In addition, when the Micro Branch and annotation refinement methods are combined, the performance improves even further, for all datasets. This confirms our hypothesis in \ref{ssec:mib_vs_refine}, and that the annotation refinement is crucial for detecting more difficult cracks.

\begin{figure*}[!tb]
    \footnotesize
    \centering
    \renewcommand{\arraystretch}{0.6}
    \setlength{\tabcolsep}{-2pt}
    \newcolumntype{B}{>{\centering\arraybackslash} m{2.1cm}@{\extracolsep{6pt}} }
    \newcolumntype{C}{>{\centering\arraybackslash} m{2.1cm} }
	\begin{tabular}{BCBCBCCC}

    \includegraphics[width=\linewidth]{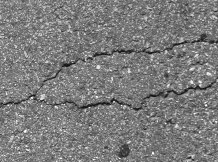} &
    \includegraphics[width=\linewidth]{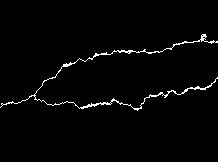} &
    \includegraphics[width=\linewidth]{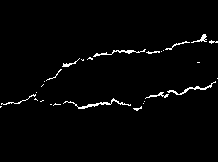} &
    \includegraphics[width=\linewidth]{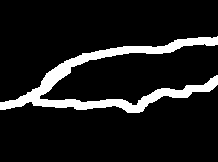} &
    \includegraphics[width=\linewidth]{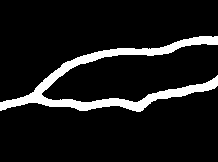} &
    \includegraphics[width=\linewidth]{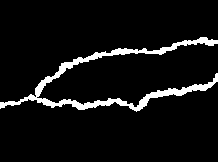} &
    \includegraphics[width=\linewidth]{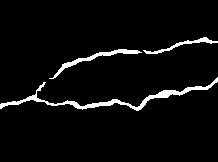} &
    \includegraphics[width=\linewidth]{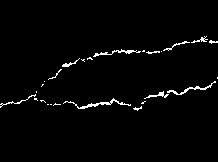} \\

    \includegraphics[width=\linewidth]{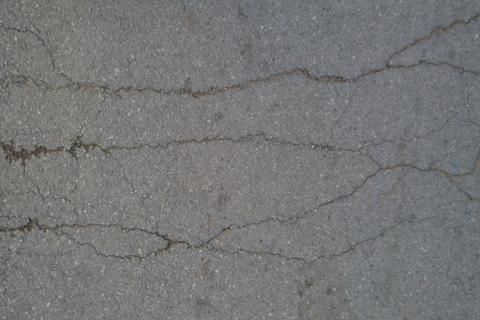} &
    \includegraphics[width=\linewidth]{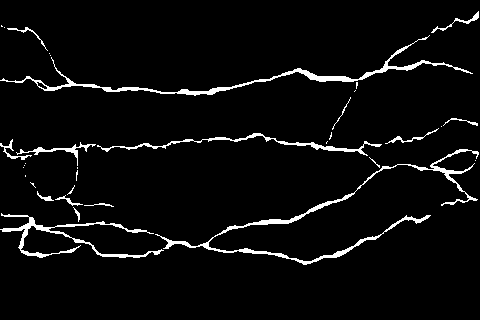} &
    \includegraphics[width=\linewidth]{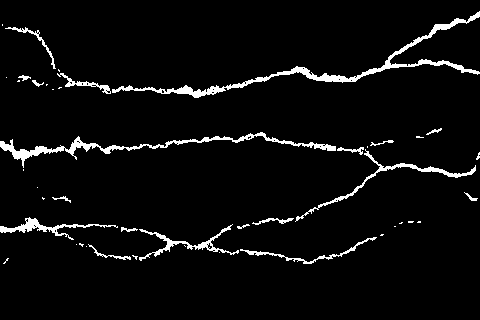} &
    \includegraphics[width=\linewidth]{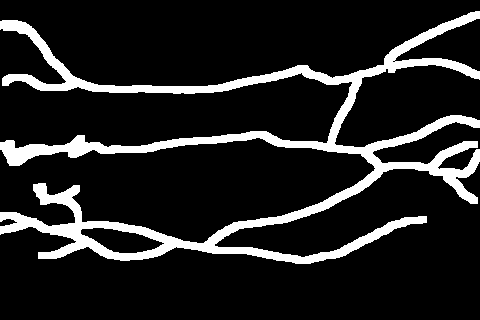} &
    \includegraphics[width=\linewidth]{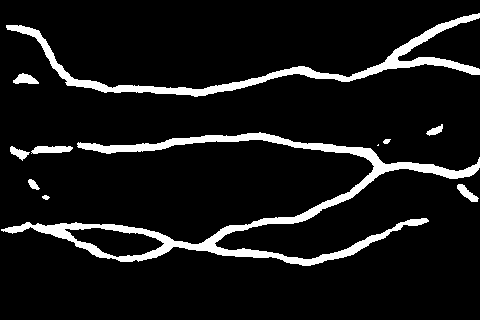} &
    \includegraphics[width=\linewidth]{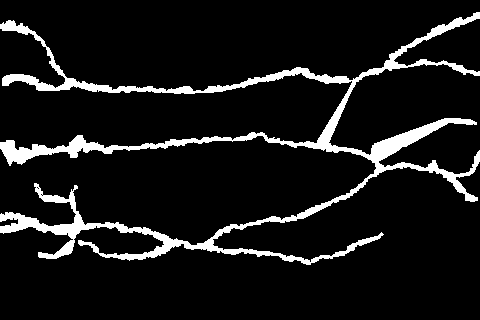} &
    \includegraphics[width=\linewidth]{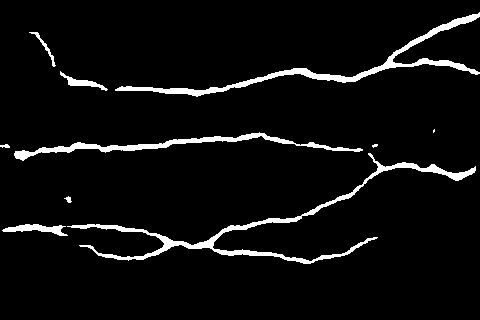} &
    \includegraphics[width=\linewidth]{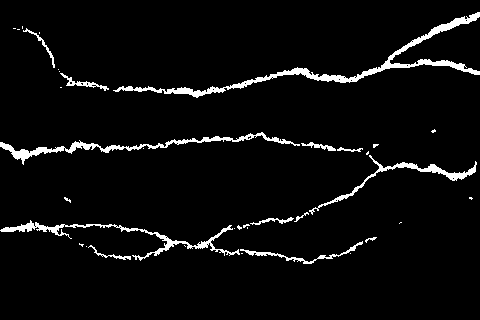} \\

    \includegraphics[width=\linewidth]{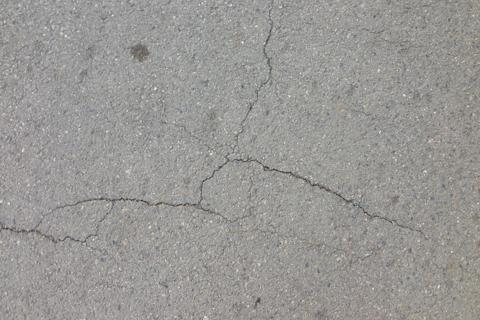} &
    \includegraphics[width=\linewidth]{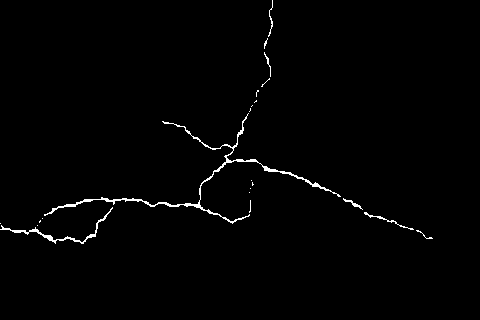} &
    \includegraphics[width=\linewidth]{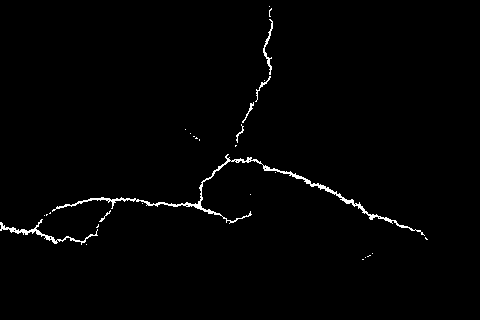} &
    \includegraphics[width=\linewidth]{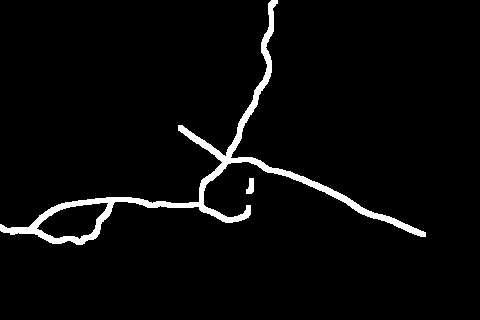} &
    \includegraphics[width=\linewidth]{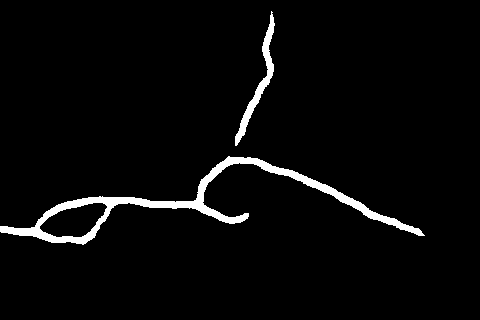} &
    \includegraphics[width=\linewidth]{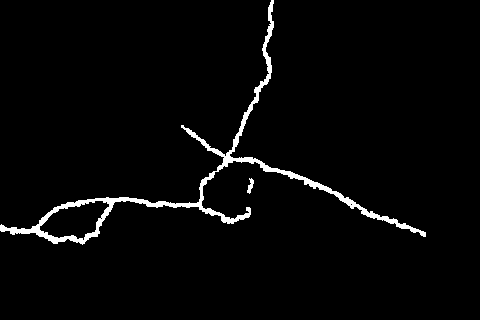} &
    \includegraphics[width=\linewidth]{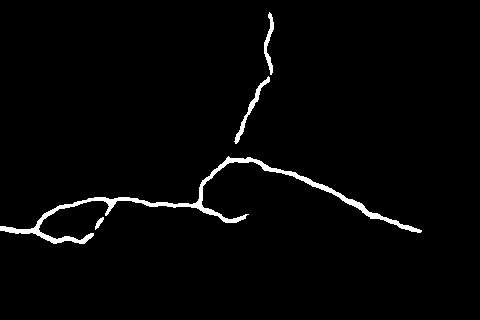} &
    \includegraphics[width=\linewidth]{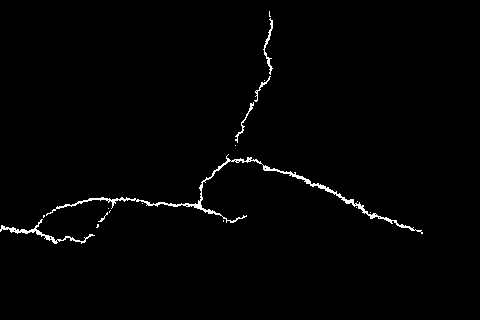} \\

    \\[-1.5mm]
    Original & Annotation & Prediction & Annotation & Prediction & Annotation & Prediction w/o MiB & Prediction \\
	\cmidrule{2-3} \cmidrule{4-5} \cmidrule{6-8}
     & \multicolumn{2}{c}{Precise Annotation} & \multicolumn{2}{c}{Rougher Annotation} & \multicolumn{3}{c}{Refined Rougher Annotation}

    \end{tabular}
    \caption{Sample predictions of the DeepCrack model under fully-supervised and weakly-supervised settings. MiB stands for the Micro Branch.}
    \label{fig:sample_inference}
\end{figure*}

Samples of annotation refinement and inference for the DeepCrack model are shown in \ref{fig:sample_inference}. As expected, models trained with less precise annotations learn to predict less precisely. The low precision output is greatly improved after the annotation is refined, and the inclusion of the Micro Branch further improves the prediction.

\subsection{Ablation Studies}

\begin{table*}[!tb]
	\centering
    \small
	\caption{Ablation studies with DeepCrack as the Macro Branch. \textbf{Bolded} and \underline{underlined} values indicate the best and second best performing settings, respectively.}
    \newcolumntype{C}{>{\centering\arraybackslash} m{0.8cm}}
    \setlength{\tabcolsep}{2pt}
	\begin{tabular}{@{\extracolsep{4pt}}lCCCCCCCCCCCC@{}}
		\toprule
		\multicolumn{1}{c}{}& \multicolumn{4}{c}{Aigle}& \multicolumn{4}{c}{CFD}& \multicolumn{4}{c}{DCD} \\
		\addlinespace[-0.5mm]
		\cmidrule{2-5} \cmidrule{6-9} \cmidrule{10-13}
		\addlinespace[-0.5mm]
		\multicolumn{1}{c}{}& R & R-er & Dil-1 & Dil-4 & R & R-er & Dil-1 & Dil-4 & R & R-er & Dil-1 & Dil-4 \\
		\addlinespace[-0.5mm]
		\midrule
        w/o Micro Branch&0.594&0.563&0.626&0.465&0.649&0.629&0.628&0.549&0.840&0.837&0.832&0.808 \\
        w/o Annotation Refinement \cite{inoue2020crack}&\textbf{0.816}&\textbf{0.808}&\underline{0.772}&\textbf{0.775}&0.629&0.590&0.636&0.574&0.835&0.830&0.831&0.823 \\
		w/o Ignore Condition&0.730&0.673&0.736&0.688&0.630&\underline{0.643}&0.637&0.595&0.842&\underline{0.846}&0.837&\underline{0.836} \\
        w/o Shrink&0.748&0.738&0.733&0.727&\underline{0.650}&0.637&\textbf{0.651}&\textbf{0.617}&\underline{0.847}&0.840&\underline{0.840}&\textbf{0.840} \\
        \midrule
        Proposed&\underline{0.778}&\underline{0.761}&\textbf{0.776}&\underline{0.765}&\textbf{0.658}&\textbf{0.644}&\underline{0.646}&\underline{0.596}&\textbf{0.849}&\textbf{0.852}&\textbf{0.844}&\underline{0.836} \\
        \bottomrule
	\end{tabular}
	\label{tbl:ablation}
\end{table*}

\ref{tbl:ablation} summarizes the ablation results. All experiments are conducted with the DeepCrack model.
The first two rows in the table correspond to \textit{Refinement only} and \textit{MiB only} plots in \ref{fig:eval_results}, respectively, and they show that (1) including Micro Branch is effective and (2) annotation refinement is effective for CFD and DCD, as discussed in \ref{ssec:eval_lq}.

\paragraph{Ignore Condition}
The ignore condition is introduced to help the model ignore mislabels by removing high crack predictions in LQ non-crack regions and low crack predictions in LQ crack regions from the loss function.
\ref{tbl:ablation} shows that the ignore condition is the most effective for Aigle. One explanation for this is that backgrounds in Aigle images have highly complex textures (\ref{fig:gt_samples} shows an example). Therefore, it should be nearly impossible for the Myopic Model to tell the difference between the complex background textures and crack regions, as it only receives highly localized input. The introduction of the ignore condition allows the Myopic Model to drop the incorrect backpropagation signals generated by the complex-textured backgrounds.

\begin{figure}[!tb]
    \centering
    \small
    \renewcommand{\arraystretch}{0.4}
    \setlength{\tabcolsep}{1pt}
    \newcolumntype{C}{>{\centering\arraybackslash} m{1.7cm}}
	\begin{tabular}{CCCCC}

    % \includegraphics[width=\linewidth,trim={0 0 1cm 0},clip]{imgs/anno_ref/Im_GT_AIGLE_RN_F11bor_output/detailed_original.png}
    % &
    % \includegraphics[width=\linewidth,trim={0 0 1cm 0},clip]{imgs/anno_ref/Im_GT_AIGLE_RN_F11bor_output/dil4_distortL0925_gt.png}
    % &
    % \includegraphics[width=\linewidth,trim={0 0 1cm 0},clip]{imgs/anno_ref/Im_GT_AIGLE_RN_F11bor_output/Im_GT_AIGLE_RN_F11bor_output.png}
    % &
    % \includegraphics[width=\linewidth,trim={0 0 1cm 0},clip]{imgs/anno_ref/Im_GT_AIGLE_RN_F11bor_output/dil4_13G1refineddil_gt.png}
    % &
    % \includegraphics[width=\linewidth,trim={0 0 1cm 0},clip]{imgs/anno_ref/Im_GT_AIGLE_RN_F11bor_output/detailed_gt.png}
    % \\

    \includegraphics[width=\linewidth,trim={0 0 1cm 0},clip,trim={0 0 1cm 0},clip]{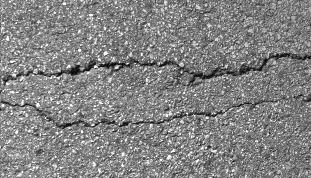}
    &
    \includegraphics[width=\linewidth,trim={0 0 1cm 0},clip]{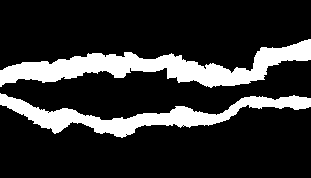}
    &
    \includegraphics[width=\linewidth,trim={0 0 1cm 0},clip]{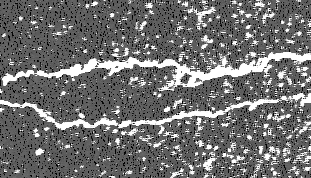}
    &
    \includegraphics[width=\linewidth,trim={0 0 1cm 0},clip]{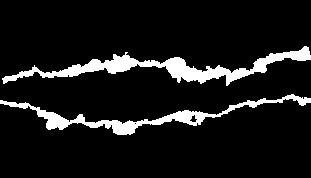}
    &
    \includegraphics[width=\linewidth,trim={0 0 1cm 0},clip]{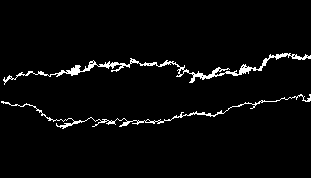}
    \\

    % \includegraphics[width=\linewidth,trim={0 0 1cm 0},clip]{imgs/anno_ref/012_output/detailed_original.png}
    % &
    % \includegraphics[width=\linewidth,trim={0 0 1cm 0},clip]{imgs/anno_ref/012_output/dil4_distortL0925_gt.png}
    % &
    % \includegraphics[width=\linewidth,trim={0 0 1cm 0},clip]{imgs/anno_ref/012_output/012_output.png}
    % &
    % \includegraphics[width=\linewidth,trim={0 0 1cm 0},clip]{imgs/anno_ref/012_output/dil4_13G1refineddil_gt.png}
    % &
    % \includegraphics[width=\linewidth,trim={0 0 1cm 0},clip]{imgs/anno_ref/012_output/detailed_gt.png}
    % \\

    \includegraphics[width=\linewidth,trim={0 0 1cm 0},clip]{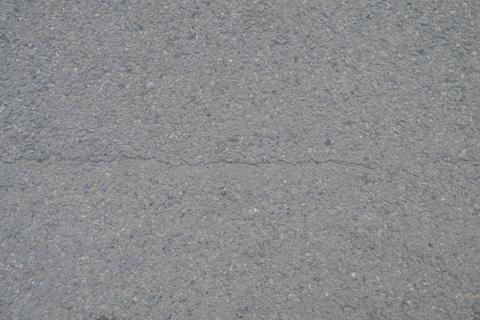}
    &
    \includegraphics[width=\linewidth,trim={0 0 1cm 0},clip]{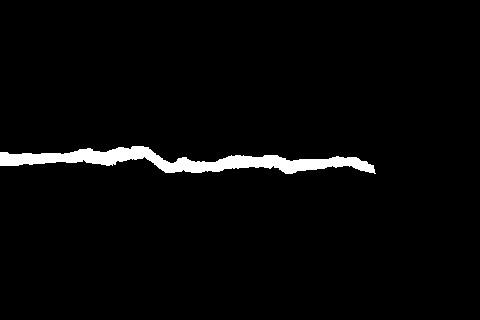}
    &
    \includegraphics[width=\linewidth,trim={0 0 1cm 0},clip]{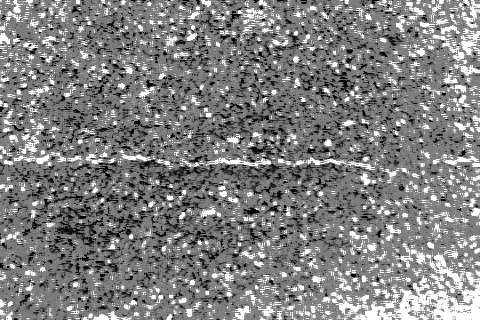}
    &
    \includegraphics[width=\linewidth,trim={0 0 1cm 0},clip]{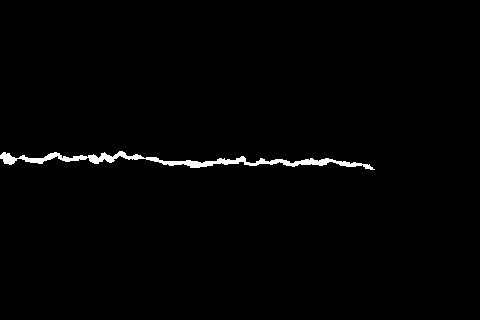}
    &
    \includegraphics[width=\linewidth,trim={0 0 1cm 0},clip]{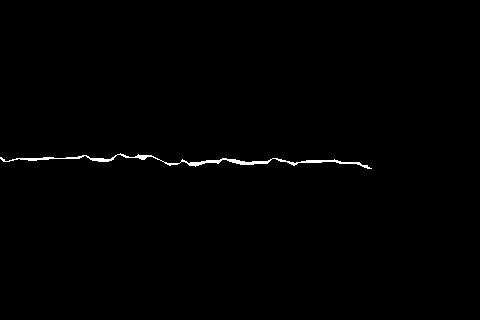}
    \\

    % \includegraphics[width=\linewidth,trim={0 0 1cm 0},clip]{imgs/anno_ref/019_output/detailed_original.png}
    % &
    % \includegraphics[width=\linewidth,trim={0 0 1cm 0},clip]{imgs/anno_ref/019_output/dil4_distortL0925_gt.png}
    % &
    % \includegraphics[width=\linewidth,trim={0 0 1cm 0},clip]{imgs/anno_ref/019_output/019_output.png}
    % &
    % \includegraphics[width=\linewidth,trim={0 0 1cm 0},clip]{imgs/anno_ref/019_output/dil4_13G1refineddil_gt.png}
    % &
    % \includegraphics[width=\linewidth,trim={0 0 1cm 0},clip]{imgs/anno_ref/019_output/detailed_gt.png}
    % \\

    % \includegraphics[width=\linewidth,trim={0 0 1cm 0},clip]{imgs/anno_ref/_11123-2_output/detailed_original.png}
    % &
    % \includegraphics[width=\linewidth,trim={0 0 1cm 0},clip]{imgs/anno_ref/_11123-2_output/dil4_distortL0925_gt.png}
    % &
    % \includegraphics[width=\linewidth,trim={0 0 1cm 0},clip]{imgs/anno_ref/_11123-2_output/_11123-2_output.png}
    % &
    % \includegraphics[width=\linewidth,trim={0 0 1cm 0},clip]{imgs/anno_ref/_11123-2_output/dil4_13G1refineddil_gt.png}
    % &
    % \includegraphics[width=\linewidth,trim={0 0 1cm 0},clip]{imgs/anno_ref/_11123-2_output/detailed_gt.png}
    % \\

    \includegraphics[width=\linewidth,trim={0 0 1cm 0},clip,trim={0 1.5cm 0 1.5cm},clip]{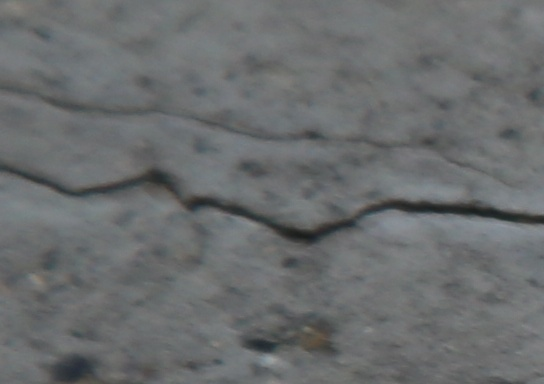}
    &
    \includegraphics[width=\linewidth,trim={0 0 1cm 0},clip,trim={0 1.5cm 0 1.5cm},clip]{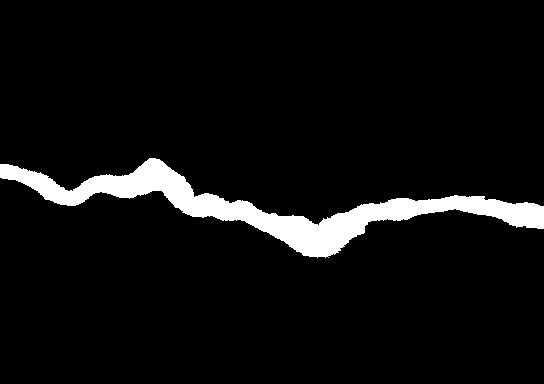}
    &
    \includegraphics[width=\linewidth,trim={0 0 1cm 0},clip,trim={0 1.5cm 0 1.5cm},clip]{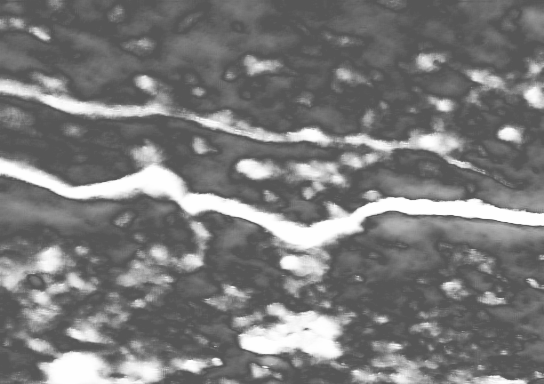}
    &
    \includegraphics[width=\linewidth,trim={0 0 1cm 0},clip,trim={0 1.5cm 0 1.5cm},clip]{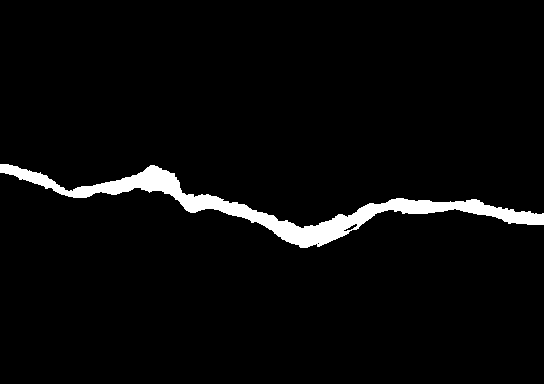}
    &
    \includegraphics[width=\linewidth,trim={0 0 1cm 0},clip,trim={0 1.5cm 0 1.5cm},clip]{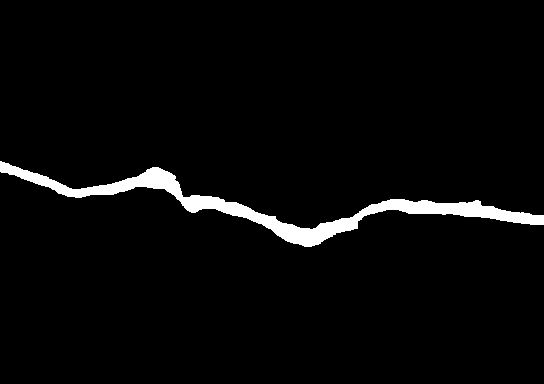}
    \\

    Input & LQ Annotation & Myopic Model & Shrink Module & Precise Annotation \\

    \end{tabular}
    \caption{Examples of the annotation refinement process.}
    \label{fig:refinement_samples}
\end{figure}

\begin{figure}[!tb]
    \centering
    \includegraphics[width=\linewidth]{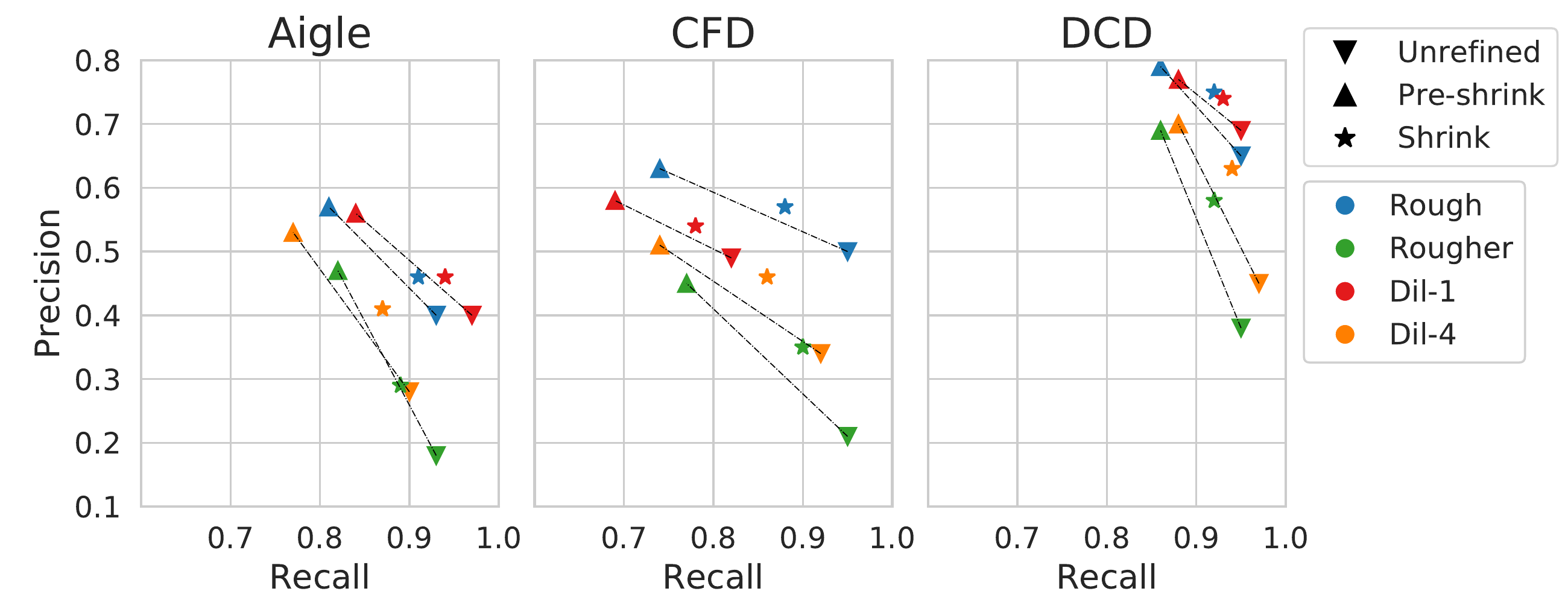}
    \caption{Annotation qualities of different refinement methods. The loss in recall introduced by the Myopic Models (\textit{Pre-shrink}) is recovered after the shrink operation, and it is better than a simple linear interpolation between Pre-shrink and Unrefined.}
    \label{fig:refinement_result}
\end{figure}

\paragraph{Shrink Module}
As shown in \ref{fig:refinement_samples}, the outputs of the Shrink Module contain fewer false-positives than the LQ annotation. In addition, they are more connected than the Myopic Model outputs and are overall more similar to the ground truth.

The effect of the annotation refinement methods on the annotation quality is also visualized in \ref{fig:refinement_result}, which plots the recall and precision of the unrefined annotation (Unrefined), the annotation in which only the Myopic Model is applied (Pre-shrink), and the annotation in which the Shrink Module is applied (Shrink), indicated by different shapes.
The figure shows that Shrink neither has the highest recall nor precision, but it always scores above the linear interpolation between Unrefined and Pre-shrink, which could indicate that Shrink achieves a better trade-off between precision and recall.

\begin{table*}[!tb]
	\centering
	\footnotesize
	
	\caption{Comparisons against various annotation reducing methods. All experiments are conducted with the DeepCrack model.}
    \newcolumntype{D}{>{\centering\arraybackslash} m{0.15cm}}
    \setlength{\tabcolsep}{2pt}
	\begin{tabular}{@{\extracolsep{4pt}}DDc@{\hskip 4mm}c@{\hskip 4mm}ccc@{\hskip 4mm}ccc@{\hskip 4mm}ccccccccc}
		\toprule
		 &&& \textit{In Domain} & \multicolumn{3}{c}{\textit{Out of Domain}}\hskip 4mm & \multicolumn{3}{c}{\textit{Transfer Learning}}\hskip 4mm  & \multicolumn{9}{c}{\textit{Weakly-Supervised}}\hskip 4mm \\
		\cmidrule(r{4mm}){4-4} \cmidrule(r{4mm}){5-7} \cmidrule(r{4mm}){8-10} \cmidrule{11-19}
		 &&& & && & &&  & \multicolumn{5}{c}{Simple Threshold} & \multicolumn{2}{c}{K\"{o}nig \etal} & \multicolumn{2}{c}{Proposed} \\
		 \cmidrule{11-15} \cmidrule{16-17} \cmidrule{18-19}
    	 && MiB & Precise & Aigle & CFD & DCD & Aigle & CFD & DCD & 0.5 & 0.4 & 0.3 & 0.2 & 0.1 & 16 & 64 & Rough & Rougher  \\
    	\midrule
	    \multirow{6}{*}{\rotatebox[origin=c]{90}{Test Dataset}} & \multirow{2}{*}{\rotatebox[origin=c]{90}{Aigle}}    &           & 0.724 &   -   & 0.534 & 0.571 &   -   & 0.554 & 0.613 & 0.591 & 0.584 & 0.674 & 0.722 & 0.755 & 0.610 & 0.613 & 0.594 & 0.563 \\
                                                            &&\checkmark & 0.787 &   -   & 0.748 & 0.694 &   -   & 0.640 & 0.659 & 0.765 & 0.756 & 0.760 & 0.762 & 0.779 & 0.641 & 0.654 & 0.778 & 0.760 \\
    	\cmidrule{2-19}
	    &\multirow{2}{*}{\rotatebox[origin=c]{90}{CFD}}      &           & 0.657 & 0.637 &   -   & 0.567 & 0.643 &   -   & 0.645 & 0.606 & 0.611 & 0.612 & 0.622 & 0.514 & 0.630 & 0.610 & 0.649 & 0.628 \\
                                                            &&\checkmark & 0.656 & 0.635 &   -   & 0.548 & 0.647 &   -   & 0.650 & 0.613 & 0.608 & 0.611 & 0.622 & 0.495 & 0.632 & 0.612 & 0.658 & 0.644 \\
    	\cmidrule{2-19}
	    &\multirow{2}{*}{\rotatebox[origin=c]{90}{DCD}}      &           & 0.841 & 0.548 & 0.330 &   -   & 0.713 & 0.700 &   -   & 0.834 & 0.837 & 0.840 & 0.828 & 0.768 & 0.768 & 0.764 & 0.840 & 0.837 \\
                                                            &&\checkmark & 0.845 & 0.565 & 0.382 &   -   & 0.734 & 0.719 &   -   & 0.837 & 0.839 & 0.841 & 0.830 & 0.769 & 0.768 & 0.765 & 0.849 & 0.852 \\
        \bottomrule
	\end{tabular}
	\label{tbl:ablation_anno_eff}

\end{table*}

\subsection{Comparisons Against Other Annotation Efficient Methods} \label{ssec:model_reuse}

In this section, we compare the proposed method against the various annotation cost reducing approaches mentioned in \ref{ssec:annotation_free}. Note that all experiments are conducted with the DeepCrack model.
The results are aggregated in \ref{tbl:ablation_anno_eff}, in which the dataset names in the leftmost column indicate the test dataset and \textit{MiB} stands for whether Micro Branch was used. \textit{In Domain} (ID) column corresponds to the fully-supervised results with accurate annotations, and the proposed method applied to the Rough and Rougher Annotations are shown in the rightmost columns.

\subsubsection{Out of Domain (OOD)}

We claimed in \ref{ssec:annotation_free} that it is more favorable to collect new sets of annotations at each site, as the appearance of cracks varies greatly. In fact, the assumption that crack detectors do not generalize well across different environments is one of the biggest motivations of this work.
The columns named \textit{Out of Domain} correspond to this setting. Each column represents the dataset from which the models are trained. For example, \textit{0.534} is the ODS performance when a model is trained on the Precise Annotation of CFD and tested on Aigle, with no Micro Branch.

The results show that compared to their ID counterparts, reusing the same model in different environments significantly degrades the performance. The drop could be anywhere between 2-46\%, even after the Micro Branch is applied. Although the table only shows the results for the DeepCrack model, a similar drop in performance was observed for all four models.
Interestingly, many models seem to generalize better when trained with Aigle, the smallest dataset. This is counter-intuitive, as larger datasets are assumed to generalize better. This makes it difficult to compile a dataset that trains robust models, as simply annotating more samples does not equate to robustness.

Comparing the OOD results with the proposed method, we can see that a small cost of annotation can lead to a significant increase in performance.
As the annotation cost of the proposed approach is kept minimal, we believe the large gain in accuracy and stability of the result compared to the OOD inference makes our proposal an appealing option when deploying crack detectors in multiple domains.

\subsubsection{Transfer Learning (TL)}

Transfer learning is similar to the OOD setting, except the models are fine-tuned with a small set of samples from the target domain.
To make the comparison fair, the annotation cost budget is kept comparable between transfer learning and that of the proposed method. To do so, the number of Precise Annotations that can be generated under the same time budget as generating Rougher Annotations for all train images is calculated based on the values summarized in \ref{tbl:dataset_overview}.
For example, there are 71 images in CFD's train set and it takes $656/22=29.8$ times longer to prepare Precise Annotation compared to Rougher Annotation. Therefore, $71/29.8=2.4$ Precise Annotations can be generated during the time it takes to generate 71 Rougher Annotations. Since the annotation time information for the Precise Annotation is only available for CFD, the same Precise to Rougher ratio (29.8) is used for all datasets, resulting in 1 image for Aigle, 3 images for CFD, and 10 images for DCD.

The columns titled \textit{Transfer Learning} correspond to this setting. As in the OOD case, each column corresponds to the datasets in which the models are pretrained. The pretrained models are fine-tuned using a randomly selected subset of the Precise Annotations from the target dataset. The size of the fine-tuning dataset varies by the target dataset, as mentioned in the previous paragraph. Furthermore, since it is likely that the model training is affected by the selection of the images to train with, a mean of five image selections is shown.

Comparing TL and OOD performances, models generally perform better after being fine-tuned as expected. One notable exception is when the models are fine-tuned and evaluated on Aigle, in which the performance drops. This is probably because the model had to fine-tune to a single image, causing over-fitting.
Interestingly, a model's performance in OOD does not seem to translate to the TL case, and the effect of the pretraining dataset on the final performance is much more subdued for TL. However, there is still a big gap between TL and in domain settings.

Comparing TL with the proposed method, we can see that the proposed method performs better for most cases, especially after MiB is applied. While TL performs better for CFD, the performance gap is minimal, especially after MiB is applied. For the other two datasets, the proposed method outperforms TL by a large margin. The largest performance gap is in DCD, which is likely because DCD contains a wide variety of target surfaces, and 10 images could not cover all cases.

\subsubsection{Simple Threshold Baseline}

In this baseline, the annotation refinement step shown in \ref{fig:overall} is replaced with a simple brightness threshold on the train image. Since the Precise Annotation is not available to drive some form of adaptive thresholding mechanism, a static threshold is used to create the refined annotation.

Columns titled \textit{Simple Threshold} correspond to this case for the Rougher Annotation, and the values below correspond to the thresholds used (all pixels darker than these values are considered as cracks). Note that the threshold only goes up to 0.5, as the brightness distribution of crack pixels is skewed toward the dark.
The results show that the best threshold setting varies across datasets. Note that comparing the performance of threshold setting as we do here requires Precise Annotation, which is not available in deployment. Therefore, choosing the best threshold setting in hindsight is not possible during real deployment.

Comparing the results against the proposed method, we can see that the proposed method outperforms except for the Aigle dataset. We believe this is caused by the strongly skewed pixel brightness distribution, as discussed in \ref{ssec:eval_lq} and shown visually in \ref{fig:histogram}. However, the simple threshold method also struggles with CFD for the same reason.

\subsubsection{K\"{o}nig \etal's Approach}

The final comparison is against the weakly-supervised method proposed by K\"{o}nig \etal, which is based on patch-level weak annotations.
In their original implementation, a classifier (ResNet50 \cite{he2016deep}) is first trained with a patch size of 128 $\times$ 128. Since each patch is given a binary label of crack or non-crack, and the patches are extracted from the Precise Annotation at 64 pixels stride, this implies that the annotation granularity is 64 pixels, as visualized in \ref{fig:weak_annot_comp}d. The trained classifier is then used to generate refined annotations, at a patch size of 32 $\times$ 32 with a stride of 16. The patch size during inference is changed from training to obtain higher-quality results.

As \ref{fig:weak_annot_comp}d shows, the patch annotation used in K\"{o}nig \etal's original implementation is significantly lower in quality and thus does not facilitate a fair comparison against our method. Therefore, their method was also evaluated with a better annotation quality. More specifically, the classifier is trained with 16 $\times$ 16 patch annotations, as visualized in \ref{fig:weak_annot_comp}e. The number 16 is selected, because the resulting annotation quality is similar to that of the Rougher Annotation, and it also matches the patch granularity K\"{o}nig \etal use to produce the refined annotation.

In the table, columns titled \textit{K\"{o}nig \etal} correspond to this case. The numbers below indicate the annotation patch size.
To our surprise, the performance stays relatively constant despite the fact that the annotation quality improved by a significant factor.
One reason for this behavior may be that because the size of the class activation map used for annotation refinement cannot be made less than 32 pixels (limited by the receptive field size of the ResNet50 architecture), there is a hard limit on its refinement capabilities.
In a way, the Myopic Model we propose can be viewed as a classifier with 3 $\times$ 3 receptive field size, going past that restriction.

Comparing the results with the proposed method, a significant performance gap is observed when the Micro Branch is included in the inference pipeline. This is because the proposed method is designed to work well with the Micro Branch, \ie the Shrink Module is designed to produce annotations with a high recall.

\subsection{Retinal Blood Vessel Segmentation}

\begin{figure*}[!tb]
    \centering
    \small
    \renewcommand{\arraystretch}{0.4}
    \setlength{\tabcolsep}{1pt}
    \newcolumntype{C}{>{\centering\arraybackslash} m{2.3cm}}
	\begin{tabular}{CCCCCC}

    \multirow{2}{*}{\includegraphics[width=\linewidth]{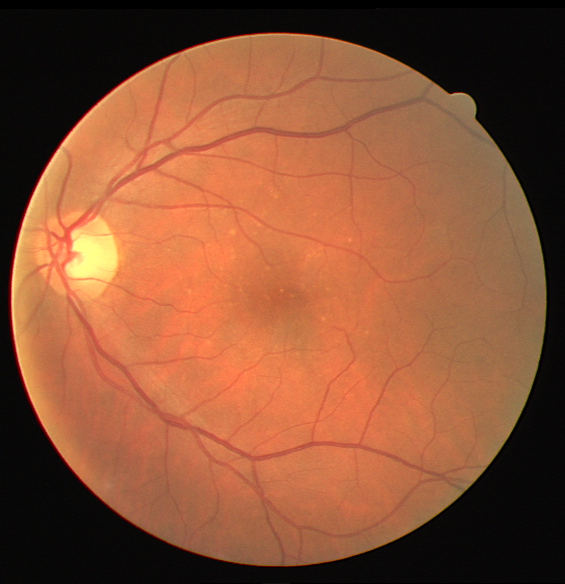}}
    &
    \includegraphics[width=\linewidth]{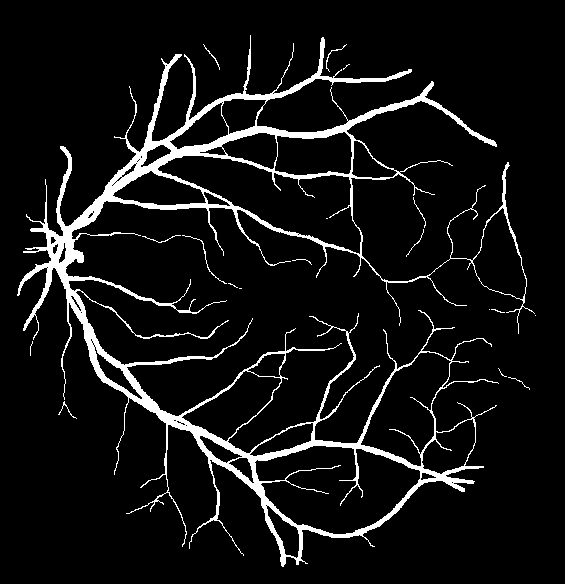}
    &
    \includegraphics[width=\linewidth]{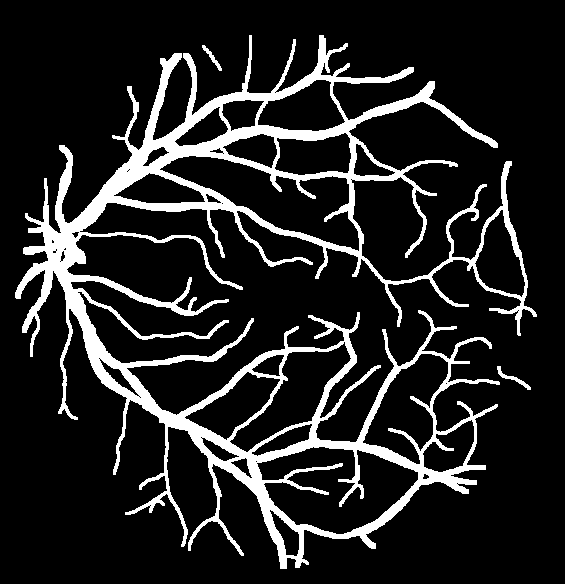}
    &
    \includegraphics[width=\linewidth]{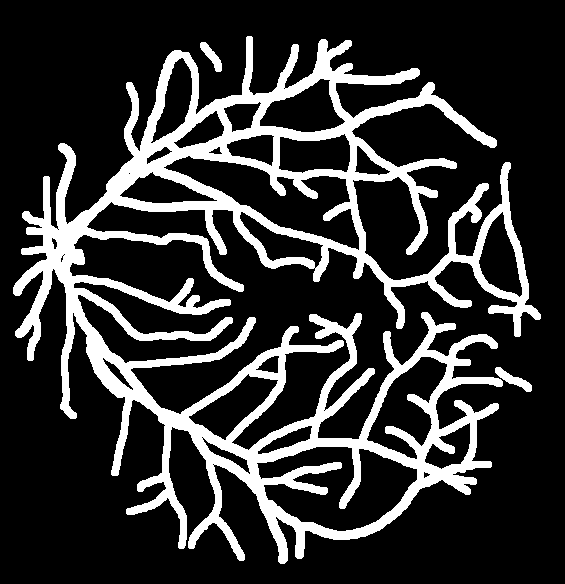}
    &
    \includegraphics[width=\linewidth]{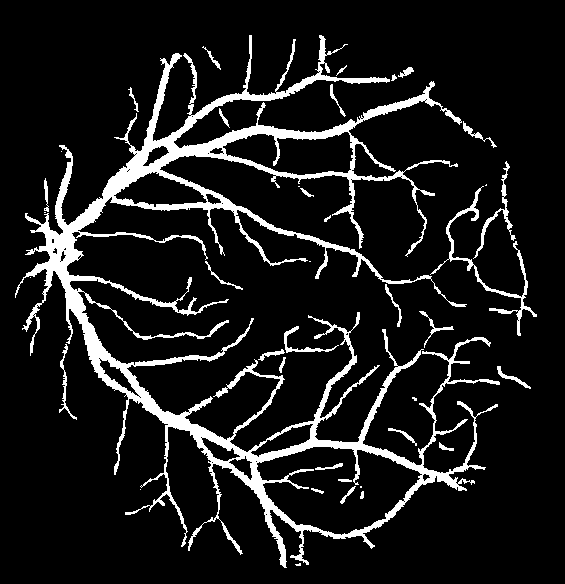}
    &
    \includegraphics[width=\linewidth]{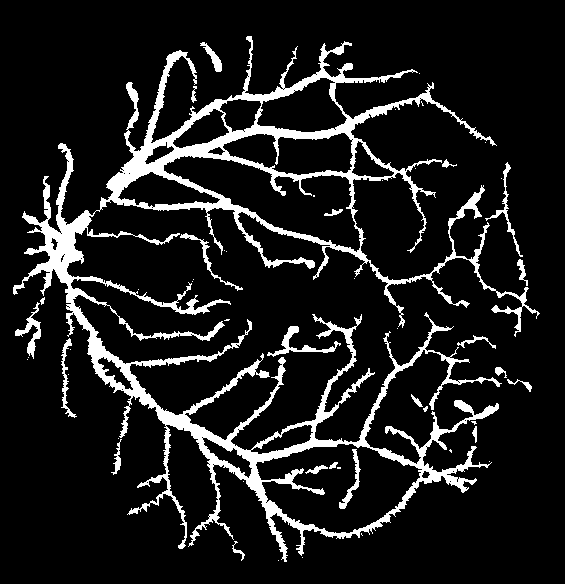}
    \\

    &
    \includegraphics[width=\linewidth]{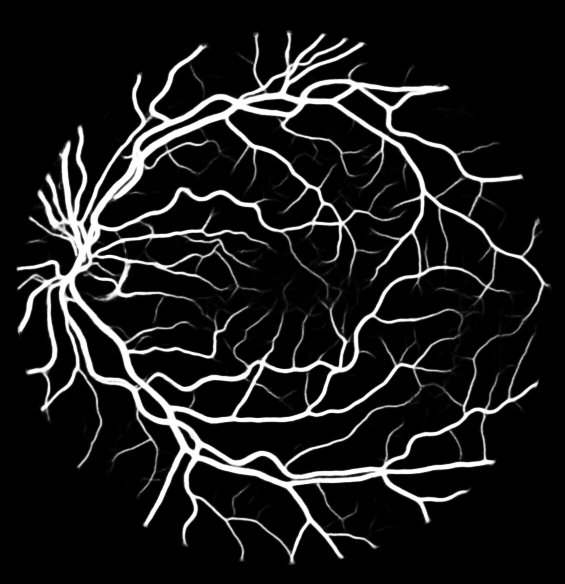}
    &
    \includegraphics[width=\linewidth]{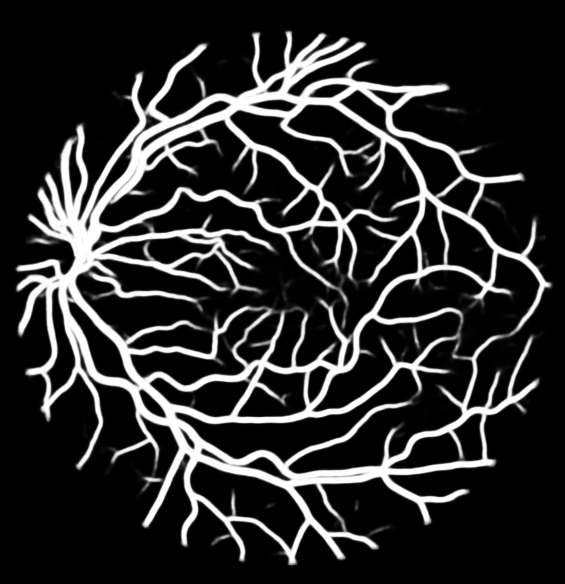}
    &
    \includegraphics[width=\linewidth]{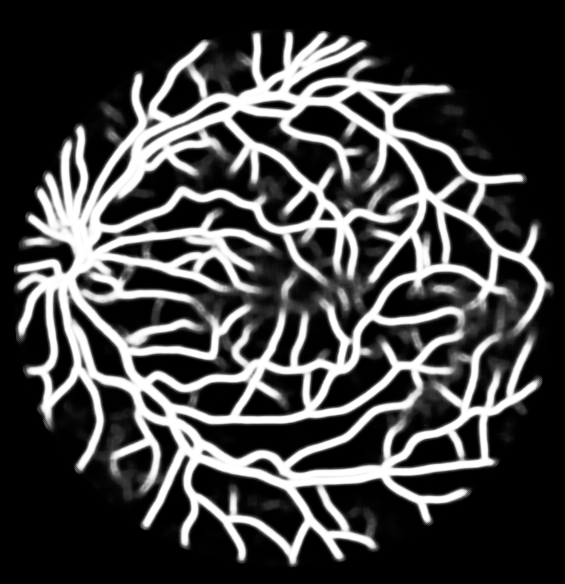}
    &
    \includegraphics[width=\linewidth]{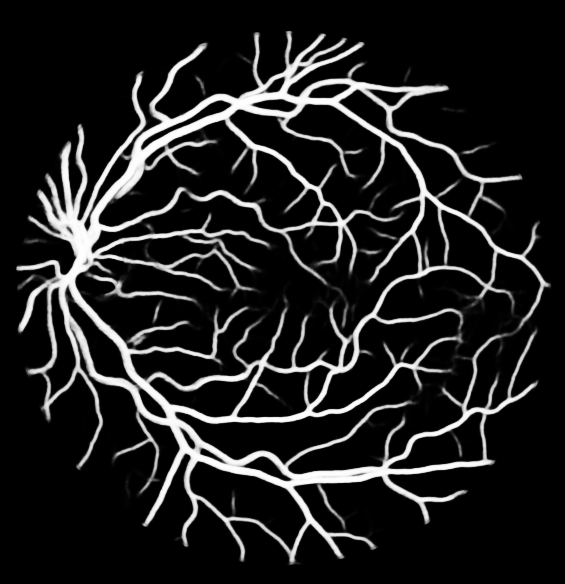}
    &
    \includegraphics[width=\linewidth]{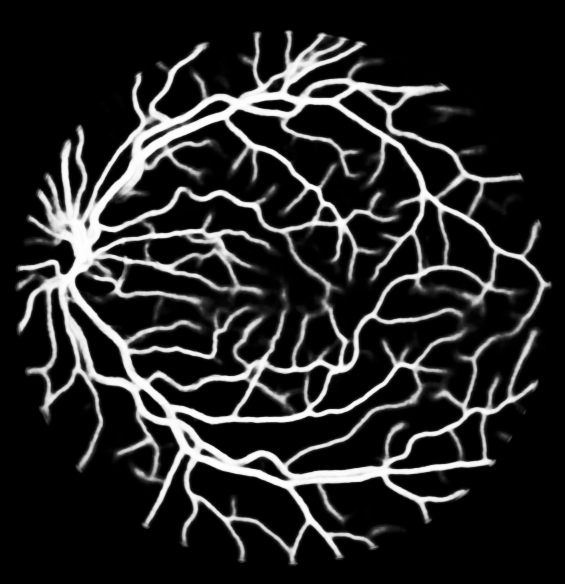}
    \\

    Input & Precise & Rough & Rougher & Rough Refined & Rougher Refined \\

    \end{tabular}
    \caption{Sample annotations (top row) and corresponding predictions (bottom row) of the DRIVE dataset.}
    \label{fig:drive_samples}
    \vspace{-3mm}
\end{figure*}

The proposed method is evaluated on the retinal blood vessel segmentation task, to see its effectiveness across different domains. SA-UNet \cite{guo2021sa} is chosen as the Macro Branch for its near-state-of-the-art performance and the availability of the code, and it is evaluated on the DRIVE dataset \cite{drive}.
Following the annotation protocols described in \ref{ssec:manual_anno}, LQ annotations of DRIVE were prepared as shown in \ref{fig:drive_samples}. For the record, it took 3.5 hours and 1.7 hours to generate the Rough and Rougher Annotations, respectively.

\begin{table}[!tb]
	\centering
	\footnotesize
	\caption{Evaluation of SA-UNet \cite{guo2021sa} on DRIVE dataset under various annotations. MiB stands for the inclusion of Micro Branch.}
    \setlength{\tabcolsep}{2pt}
	\begin{tabular}{@{\extracolsep{4pt}}lccccc}
	    \toprule
        Annotation &MiB& Refined   & Sensitivity & Specificity & F1 \\
        \midrule
        Precise &           &           & 0.808 & 0.985 & 0.821 \\
		\midrule
        \multirow{4}{*}{Rough}&           &           & 0.941 & 0.927 & 0.696 \\
                & \checkmark&           & 0.653 & 0.977 & 0.691 \\
                &           & \checkmark& 0.889 & 0.962 & 0.780\\
                & \checkmark& \checkmark& 0.564 & 0.990 & 0.678 \\
		\midrule
        \multirow{4}{*}{Rougher}&           &           & 0.958 & 0.852 & 0.547 \\
                & \checkmark&           & 0.669 & 0.954 & 0.623 \\
                &           & \checkmark& 0.910 & 0.948 & 0.742 \\
                & \checkmark& \checkmark& 0.576 & 0.988 & 0.677 \\
        \bottomrule
	\end{tabular}
	\label{tbl:drive_results}
    \vspace{-2mm}
\end{table}

\ref{tbl:drive_results} summarizes the results.
Note that the evaluation metrics are in accordance with the retinal blood vessel segmentation literature. For example, a single threshold value of 0.5 is used for evaluation, unlike the adaptive thresholding evaluation used in ODS.
As the F1-score shows, the annotation refinement is effective for retinal blood vessel segmentation as well. This can also be observed qualitatively in \ref{fig:drive_samples}.
On the other hand, the Micro Branch performed poorly for DRIVE. This can be explained by the fact that the brightness distributions of DRIVE are strongly overlapping between blood vessels and background as shown in \ref{fig:histogram}.
This confirms the remarks made in \ref{ssec:mib_vs_refine}, which is that because the annotation refinement is data-driven and thus makes fewer assumptions about the characteristics of the segmentation target, it is more robust against changes in domains.
A similar observation was made in \ref{ssec:eval_lq} for CFD, but the difference in performance between the Micro Branch and annotation refinement is larger for DRIVE; adding Micro Branch to annotation refinement actually hurts the performance for DRIVE.
This indicates that depending on the brightness distribution of the segmentation target, the Micro Branch may need to be removed from the inference path.

\section{Conclusion}

In this paper, we reformulated crack segmentation as a weakly-supervised problem to reduce the annotation time, and proposed an annotation refinement and an augmented inference framework to counteract the loss in annotation quality.
The effectiveness of the proposed framework was shown empirically- it was able to maintain accuracy even when the annotation quality was undermined. Furthermore, it consistently outperformed the various annotation-reducing methods, confirming the superiority of our proposed framework. Finally, the experiments show that the proposed framework can even be applied to non-crack targets.

\section*{Acknowledgments}
We would like to thank Hiroki Ohashi for helpful discussions during paper writing.

% {\appendix[Proof of the Zonklar Equations]
% Use $\backslash${\tt{appendix}} if you have a single appendix:
% Do not use $\backslash${\tt{section}} anymore after $\backslash${\tt{appendix}}, only $\backslash${\tt{section*}}.
% If you have multiple appendixes use $\backslash${\tt{appendices}} then use $\backslash${\tt{section}} to start each appendix.
% You must declare a $\backslash${\tt{section}} before using any $\backslash${\tt{subsection}} or using $\backslash${\tt{label}} ($\backslash${\tt{appendices}} by itself
%  starts a section numbered zero.)}

\appendices{

\section{Results on the Weakly-Supervised Annotations}

\begin{table*}
	\centering
	\caption{ODS for training with the Synthetic Annotations. Integers below model names correspond to $n_{dil}$, the number of dilation operations applied during the data synthesis process. AR and MiB stand for Annotation Refinement and Micro Branch, respectively.}

    \newcolumntype{B}{>{\centering\arraybackslash} m{0cm} }
    \newcolumntype{D}{>{\centering\arraybackslash} m{0.35cm}}
    \newcolumntype{C}{>{\centering\arraybackslash} m{0.55cm}}
    \setlength{\tabcolsep}{1.5pt}
	\begin{tabular}{@{\extracolsep{4pt}}D|cc|CCCCCCCCCCCCCCCCCCCC@{}}
		\toprule
		 &&& \multicolumn{5}{c}{MIL-CD Light} & \multicolumn{5}{c}{MIL-CD} & \multicolumn{5}{c}{DeepCrack} & \multicolumn{5}{c}{DeepLab V3+} \\
		\cline{4-8} \cline{9-13} \cline{14-18} \cline{19-23}
    	 &AR&MiB&0 &1 &2 &3 &4 &0 &1 &2 &3 &4 &0 &1 &2 &3 &4 &0 &1 &2 &3 &4   \\
    	\midrule

	    \multirow{4}{*}{\rotatebox[origin=c]{90}{Aigle}}  &           &           & 0.477&0.501&0.434&0.419&0.416&0.725&0.516&0.468&0.407&0.366&0.724&0.579&0.483&0.411&0.323&0.616&0.511&0.465&0.409&0.408  \\
                                &           &\checkmark & 0.691&0.788&0.787&0.776&0.768&0.803&0.806&0.800&0.796&0.775&0.787&0.772&0.781&0.789&0.774&0.806&0.794&0.787&0.792&0.765  \\
                                &\checkmark &           & 0.477&0.521&0.475&0.486&0.485&0.725&0.536&0.533&0.495&0.461&0.724&0.625&0.555&0.537&0.465&0.616&0.567&0.535&0.492&0.476  \\
                                &\checkmark &\checkmark & 0.691&0.786&0.795&0.788&0.772&0.803&0.802&0.804&0.804&0.796&0.787&0.776&0.780&0.765&0.765&0.806&0.830&0.814&0.808&0.782  \\
        \midrule
        
        \multirow{4}{*}{\rotatebox[origin=c]{90}{CFD}}&           &           & 0.587&0.602&0.557&0.517&0.498&0.678&0.627&0.583&0.555&0.533&0.657&0.603&0.54&0.501&0.445&0.667&0.635&0.516&0.431&0.372  \\
                            &           &\checkmark & 0.623&0.610&0.586&0.567&0.557&0.670&0.615&0.588&0.572&0.558&0.656&0.636&0.601&0.588&0.574&0.666&0.628&0.592&0.575&0.563  \\
                            &\checkmark &           & 0.587&0.625&0.611&0.588&0.580&0.678&0.644&0.640&0.622&0.611&0.657&0.628&0.615&0.595&0.549&0.667&0.661&0.627&0.590&0.553  \\
                            &\checkmark &\checkmark & 0.623&0.623&0.609&0.601&0.587&0.670&0.628&0.619&0.609&0.598&0.656&0.646&0.633&0.625&0.596&0.666&0.647&0.621&0.604&0.589  \\
        \midrule
        
        \multirow{4}{*}{\rotatebox[origin=c]{90}{DCD}}&           &           & 0.782&0.756&0.699&0.698&0.640&0.795&0.783&0.734&0.709&0.674&0.841&0.799&0.776&0.769&0.749&0.842&0.786&0.764&0.685&0.642  \\
                            &           &\checkmark & 0.824&0.821&0.813&0.810&0.805&0.822&0.831&0.822&0.817&0.813&0.845&0.831&0.824&0.833&0.823&0.848&0.818&0.825&0.817&0.797  \\
                            &\checkmark &           & 0.782&0.790&0.777&0.772&0.766&0.795&0.802&0.797&0.784&0.763&0.841&0.832&0.822&0.815&0.808&0.842&0.820&0.809&0.794&0.764  \\
                            &\checkmark &\checkmark & 0.824&0.823&0.830&0.830&0.826&0.822&0.833&0.833&0.830&0.827&0.845&0.844&0.844&0.840&0.836&0.848&0.831&0.830&0.818&0.810  \\
        \bottomrule
	\end{tabular}
	\label{tbl:table_dil_result}
\end{table*}

\begin{table*}
	\centering
	\caption{ODS for training with the Rough and Rougher Annotations.}

    \newcolumntype{D}{>{\centering\arraybackslash} m{0.35cm}}
    \newcolumntype{C}{>{\centering\arraybackslash} m{1cm}}
    \setlength{\tabcolsep}{2pt}
	\begin{tabular}{@{\extracolsep{4pt}}D|cc|CCCCCCCCCCCC@{}}
		\toprule
		 &&& \multicolumn{3}{c}{MIL-CD Light} & \multicolumn{3}{c}{MIL-CD} & \multicolumn{3}{c}{DeepCrack} & \multicolumn{3}{c}{DeepLab V3+} \\
		\cline{4-6} \cline{7-9} \cline{10-12} \cline{13-15}
    	 &AR&MiB&Precise &Rough &Rougher &Precise &Rough &Rougher &Precise &Rough &Rougher &Precise &Rough &Rougher  \\
    	\midrule
	    \multirow{4}{*}{\rotatebox[origin=c]{90}{Aigle}}  &           &           & 0.477&0.433&0.383&0.725&0.522&0.435&0.724&0.538&0.410&0.616&0.487&0.407  \\
                                &           &\checkmark & 0.691&0.773&0.781&0.803&0.816&0.802&0.787&0.816&0.808&0.806&0.773&0.811  \\
                                &\checkmark &           & 0.477&0.548&0.493&0.725&0.522&0.534&0.724&0.594&0.563&0.616&0.519&0.559  \\
                                &\checkmark &\checkmark & 0.691&0.783&0.782&0.803&0.800&0.809&0.787&0.778&0.760&0.806&0.815&0.804  \\
        \midrule
        
        \multirow{4}{*}{\rotatebox[origin=c]{90}{CFD}}&           &           & 0.587&0.586&0.519&0.678&0.625&0.588&0.657&0.624&0.536&0.667&0.649&0.506  \\
                            &           &\checkmark & 0.623&0.621&0.591&0.670&0.631&0.597&0.656&0.629&0.590&0.666&0.641&0.599  \\
                            &\checkmark &           & 0.587&0.633&0.598&0.678&0.644&0.632&0.657&0.649&0.628&0.667&0.649&0.629  \\
                            &\checkmark &\checkmark & 0.623&0.629&0.611&0.670&0.641&0.621&0.656&0.658&0.644&0.666&0.643&0.622  \\
        \midrule
        
        \multirow{4}{*}{\rotatebox[origin=c]{90}{DCD}}&           &           & 0.782&0.770&0.701&0.795&0.796&0.722&0.841&0.813&0.794&0.842&0.825&0.710  \\
                            &           &\checkmark & 0.824&0.824&0.811&0.822&0.836&0.813&0.845&0.835&0.830&0.848&0.842&0.823  \\
                            &\checkmark &           & 0.782&0.799&0.781&0.795&0.806&0.774&0.841&0.840&0.837&0.842&0.824&0.825  \\
                            &\checkmark &\checkmark & 0.824&0.831&0.818&0.822&0.837&0.833&0.845&0.849&0.852&0.848&0.843&0.833  \\
        \bottomrule
	\end{tabular}
	\label{tbl:table_man_result}
\end{table*}

Tabular versions of \ref{fig:eval_results} are provided in \ref{tbl:table_dil_result} and \ref{tbl:table_man_result} for precise numerical comparisons.

\section{Test-Train Split for Aigle and CFD}

Although both Aigle and CFD are well-known public datasets used in many crack detection literature, we were not able to find any official test-train split for those datasets. So we record our split in this section to promote future research.

\paragraph{Aigle Test Data}
C18bor, E17aor, E17bor, F01aor, F02aor, F04bor, F05bor, F08bor, F09aor, F10bor, F12bor, F13aor, F14aor, F16aor.

\paragraph{CFD Test Data}
002, 004, 005, 006, 014, 016, 018, 024, 025, 027, 028, 029, 033, 036, 037, 038, 041, 044, 047, 049, 053, 059, 060, 062, 064, 066, 073, 074, 076, 077, 078, 085, 090, 091, 093, 094, 096, 098, 102, 104, 108, 110, 111, 112, 114, 116, 118.
}


\bibliographystyle{IEEEtran}
\bibliography{IEEEabrv,egbib}

\begin{thebibliography}{10}
\providecommand{\url}[1]{#1}
\csname url@samestyle\endcsname
\providecommand{\newblock}{\relax}
\providecommand{\bibinfo}[2]{#2}
\providecommand{\BIBentrySTDinterwordspacing}{\spaceskip=0pt\relax}
\providecommand{\BIBentryALTinterwordstretchfactor}{4}
\providecommand{\BIBentryALTinterwordspacing}{\spaceskip=\fontdimen2\font plus
\BIBentryALTinterwordstretchfactor\fontdimen3\font minus
  \fontdimen4\font\relax}
\providecommand{\BIBforeignlanguage}[2]{{%
\expandafter\ifx\csname l@#1\endcsname\relax
\typeout{** WARNING: IEEEtran.bst: No hyphenation pattern has been}%
\typeout{** loaded for the language `#1'. Using the pattern for}%
\typeout{** the default language instead.}%
\else
\language=\csname l@#1\endcsname
\fi
#2}}
\providecommand{\BIBdecl}{\relax}
\BIBdecl

\bibitem{inoue}
Y.~Inoue and H.~Nagayoshi, ``Deployment conscious automatic surface crack
  detection,'' in \emph{WACV}.\hskip 1em plus 0.5em minus 0.4em\relax IEEE,
  2019, pp. 686--694.

\bibitem{yang2019feature}
F.~Yang, L.~Zhang, S.~Yu, D.~Prokhorov, X.~Mei, and H.~Ling, ``Feature pyramid
  and hierarchical boosting network for pavement crack detection,'' \emph{IEEE
  Transactions on Intelligent Transportation Systems}, 2019.

\bibitem{liu2019deepcrack}
Y.~Liu, J.~Yao, X.~Lu, R.~Xie, and L.~Li, ``Deepcrack: A deep hierarchical
  feature learning architecture for crack segmentation,''
  \emph{Neurocomputing}, vol. 338, pp. 139--153, 2019.

\bibitem{guo2021barnet}
J.-M. Guo, H.~Markoni, and J.-D. Lee, ``Barnet: Boundary aware refinement
  network for crack detection,'' \emph{IEEE Transactions on Intelligent
  Transportation Systems}, 2021.

\bibitem{qu2021crack}
Z.~Qu, W.~Chen, S.-Y. Wang, T.-M. Yi, and L.~Liu, ``A crack detection algorithm
  for concrete pavement based on attention mechanism and multi-features
  fusion,'' \emph{IEEE Transactions on Intelligent Transportation Systems},
  2021.

\bibitem{inoue2020crack}
Y.~Inoue and H.~Nagayoshi, ``Crack detection as a weakly-supervised problem:
  Towards achieving less annotation-intensive crack detectors,'' in
  \emph{ICPR}, 2020.

\bibitem{konig2021weakly}
J.~K{\"o}nig, M.~D. Jenkins, M.~Mannion, P.~Barrie, and G.~Morison,
  ``Weakly-supervised surface crack segmentation by generating pseudo-labels
  using localization with a classifier and thresholding,'' \emph{IEEE
  Transactions on Intelligent Transportation Systems}, 2022.

\bibitem{mubashshira2020unsupervised}
S.~Mubashshira, M.~M. Azam, and S.~M.~M. Ahsan, ``An unsupervised approach for
  road surface crack detection,'' in \emph{2020 IEEE Region 10 Symposium
  (TENSYMP)}.\hskip 1em plus 0.5em minus 0.4em\relax IEEE, 2020, pp.
  1596--1599.

\bibitem{duan2020unsupervised}
L.~Duan, H.~Geng, J.~Pang, and J.~Zeng, ``Unsupervised pixel-level crack
  detection based on generative adversarial network,'' in \emph{Proceedings of
  the 2020 5th International Conference on Multimedia Systems and Signal
  Processing}, 2020, pp. 6--10.

\bibitem{yu2020unsupervised}
J.~Yu, D.~Y. Kim, Y.~Lee, and M.~Jeon, ``Unsupervised pixel-level road defect
  detection via adversarial image-to-frequency transform,'' in \emph{2020 IEEE
  Intelligent Vehicles Symposium (IV)}.\hskip 1em plus 0.5em minus 0.4em\relax
  IEEE, 2020, pp. 1708--1713.

\bibitem{fan2019road}
R.~Fan, M.~J. Bocus, Y.~Zhu, J.~Jiao, L.~Wang, F.~Ma, S.~Cheng, and M.~Liu,
  ``Road crack detection using deep convolutional neural network and adaptive
  thresholding,'' in \emph{IEEE Intelligent Vehicles Symposium}.\hskip 1em plus
  0.5em minus 0.4em\relax IEEE, 2019, pp. 474--479.

\bibitem{dong2020patch}
Z.~Dong, J.~Wang, B.~Cui, D.~Wang, and X.~Wang, ``Patch-based weakly supervised
  semantic segmentation network for crack detection,'' \emph{Construction and
  Building Materials}, vol. 258, p. 120291, 2020.

\bibitem{selvaraju2017grad}
R.~R. Selvaraju, M.~Cogswell, A.~Das, R.~Vedantam, D.~Parikh, and D.~Batra,
  ``Grad-cam: Visual explanations from deep networks via gradient-based
  localization,'' in \emph{CVPR}, 2017, pp. 618--626.

\bibitem{li2020semi}
G.~Li, J.~Wan, S.~He, Q.~Liu, and B.~Ma, ``Semi-supervised semantic
  segmentation using adversarial learning for pavement crack detection,''
  \emph{IEEE Access}, vol.~8, pp. 51\,446--51\,459, 2020.

\bibitem{shim2020multiscale}
S.~Shim, J.~Kim, G.-C. Cho, and S.-W. Lee, ``Multiscale and adversarial
  learning-based semi-supervised semantic segmentation approach for crack
  detection in concrete structures,'' \emph{IEEE Access}, vol.~8, pp.
  170\,939--170\,950, 2020.

\bibitem{opencv_library}
G.~Bradski, ``{The OpenCV Library},'' \emph{Dr. Dobb's Journal of Software
  Tools}, 2000.

\bibitem{scikit-image}
\BIBentryALTinterwordspacing
S.~van~der Walt, J.~L. {S}ch\"onberger, J.~{Nunez-Iglesias}, F.~{B}oulogne,
  J.~D. {W}arner, N.~{Y}ager, E.~{G}ouillart, T.~{Y}u, and the scikit-image
  contributors, ``scikit-image: image processing in {P}ython,'' \emph{PeerJ},
  vol.~2, p. e453, 6 2014. [Online]. Available:
  \url{https://doi.org/10.7717/peerj.453}
\BIBentrySTDinterwordspacing

\bibitem{aiglern}
S.~Chambon and J.-M. Moliard, ``Automatic road pavement assessment with image
  processing: review and comparison,'' \emph{International Journal of
  Geophysics}, vol. 2011, 2011.

\bibitem{cfd}
Y.~Shi, L.~Cui, Z.~Qi, F.~Meng, and Z.~Chen, ``Automatic road crack detection
  using random structured forests,'' \emph{IEEE Transactions on Intelligent
  Transportation Systems}, vol.~17, no.~12, pp. 3434--3445, 2016.

\bibitem{huang2022nha12d}
Z.~Huang, A.~Al-Tabbaa, I.~Brilakis \emph{et~al.}, ``Nha12d: A new pavement
  crack dataset and a comparison study of crack detection algorithms,'' 2022.

\bibitem{simard2003best}
P.~Y. Simard, D.~Steinkraus, J.~C. Platt \emph{et~al.}, ``Best practices for
  convolutional neural networks applied to visual document analysis.'' in
  \emph{ICDAR}, vol.~3, 2003.

\bibitem{buslaev2020albumentations}
A.~Buslaev, V.~I. Iglovikov, E.~Khvedchenya, A.~Parinov, M.~Druzhinin, and
  A.~A. Kalinin, ``Albumentations: fast and flexible image augmentations,''
  \emph{Information}, vol.~11, no.~2, p. 125, 2020.

\bibitem{deeplabv3p}
L.-C. Chen, Y.~Zhu, G.~Papandreou, F.~Schroff, and H.~Adam, ``Encoder-decoder
  with atrous separable convolution for semantic image segmentation,'' in
  \emph{ECCV}, 2018.

\bibitem{he2016deep}
K.~He, X.~Zhang, S.~Ren, and J.~Sun, ``Deep residual learning for image
  recognition,'' in \emph{CVPR}, 2016, pp. 770--778.

\bibitem{guo2021sa}
C.~Guo, M.~Szemenyei, Y.~Yi, W.~Wang, B.~Chen, and C.~Fan, ``Sa-unet: Spatial
  attention u-net for retinal vessel segmentation,'' in \emph{ICPR}.\hskip 1em
  plus 0.5em minus 0.4em\relax IEEE, 2021, pp. 1236--1242.

\bibitem{drive}
J.~Staal, M.~D. Abr{\`a}moff, M.~Niemeijer, M.~A. Viergever, and
  B.~Van~Ginneken, ``Ridge-based vessel segmentation in color images of the
  retina,'' \emph{IEEE transactions on medical imaging}, vol.~23, no.~4, pp.
  501--509, 2004.

\end{thebibliography}

\newpage

\section*{Biography Section}

\begin{IEEEbiography}[{\includegraphics[width=1in,height=1.25in,clip,keepaspectratio]{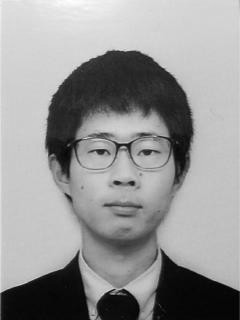}}]{Yuki Inoue}
received the B.S. and M.S. degrees in Electrical Engineering from Stanford University, CA, USA, in 2016 and 2017, respectively. He currently works for Hitachi Ltd., Tokyo, Japan as a researcher in the field of computer vision.
\end{IEEEbiography}

\begin{IEEEbiography}[{\includegraphics[width=1in,height=1.25in,clip,keepaspectratio]{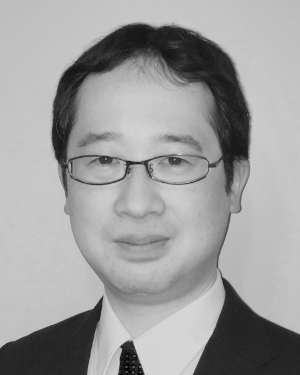}}]{Hiroto Nagayoshi}
received the B.E. and M.E. degrees in Electrical Engineering from Waseda University, Tokyo, Japan, in 1999 and 2001, respectively. Since 2001, he has worked in the field of pattern recognition and image recognition in Hitachi, Ltd., Tokyo, Japan. He was a visiting researcher at ETH, Zurich for one year from 2013 to 2014.
He was a recipient of the 55th Okouchi Memorial Technology Award, the Excellence Prize of the 3rd Monodzukuri Nippon Grand Award, and the Encouragement Prize by Minister of Education and Science of 2011 Chubu Region Invention Award.
\end{IEEEbiography}

\end{document}